# Macro-FF: Improving AI Planning with Automatically Learned Macro-Operators

**Adi Botea**                                                                 adib@cs.ualberta.ca
**Markus Enzenberger**                                                  emarkus@cs.ualberta.ca
**Martin Müller**                                                         mmueller@cs.ualberta.ca
**Jonathan Schaeffer**                                                  jonathan@cs.ualberta.ca
*Department of Computing Science, University of Alberta*
*Edmonton, AB Canada T6G 2E8*

## Abstract

Despite recent progress in AI planning, many benchmarks remain challenging for current planners. In many domains, the performance of a planner can greatly be improved by discovering and exploiting information about the domain structure that is not explicitly encoded in the initial PDDL formulation. In this paper we present and compare two automated methods that learn relevant information from previous experience in a domain and use it to solve new problem instances. Our methods share a common four-step strategy. First, a domain is analyzed and structural information is extracted, then macro-operators are generated based on the previously discovered structure. A filtering and ranking procedure selects the most useful macro-operators. Finally, the selected macros are used to speed up future searches.

We have successfully used such an approach in the fourth international planning competition IPC-4. Our system, Macro-FF, extends Hoffmann's state-of-the-art planner FF 2.3 with support for two kinds of macro-operators, and with engineering enhancements. We demonstrate the effectiveness of our ideas on benchmarks from international planning competitions. Our results indicate a large reduction in search effort in those complex domains where structural information can be inferred.

## 1. Introduction

AI planning has recently made great advances. The evolution of the international planning competition over its four editions (Bacchus, 2001; Hoffmann, Edelkamp, Englert, Liporace, Thiébaux, & Trüg, 2004; Long & Fox, 2003; McDermott, 2000) accurately reflects this. Successive editions introduced more and more complex and realistic benchmarks, or harder problem instances in the same domain. The top performers could successfully solve a large percentage of the problems each time. However, many hard domains, including benchmarks used in IPC-4, still pose great challenges for current automated planning systems.

The main claim of this paper is that in many domains, the performance of a planner can be improved by inferring and exploiting information about the domain structure that is not explicitly encoded in the initial PDDL formulation. The implicit structural information that a domain encodes is, arguably, proportional to how complex the domain is, and how realistically this models the world. For example, consider driving a truck between two locations. This operation is composed of many subtasks in the real world. To name just a few, the truck should be fueled and have a driver assigned. In a detailed planning formulation, we would define several operators such as FUEL, ASSIGN-DRIVER, and DRIVE. This representation already contains implicit information about the domain structure. It is quite obvious for a human that driving a truck between two remote locations would be a macro-action where we first fuel the truck and assign a driver (with no ordering constraints between these two actions) and next we apply the drive operator. In a simpler formulation, we can remove the operators FUEL and ASSIGN-DRIVER and consider that, in our model, a truck needs neither fuel nor a driver. Now driving a truck is modeled as a single action, and the details described above are removed from the model.

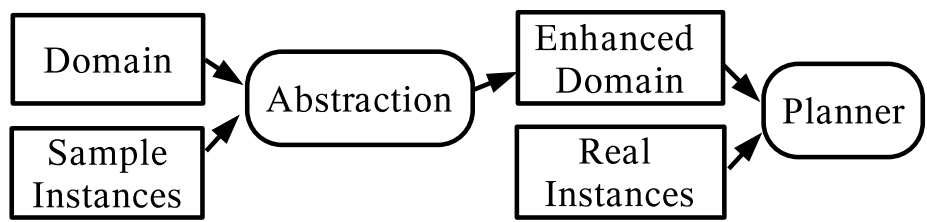

Figure 1: CA-ED – Integrating component abstraction and macro-operators into a standard planning framework.

In this article we present and evaluate two automated methods that learn such implicit domain knowledge and use it to simplify planning for new problem instances. The learning uses several training problems from a domain. Our methods share a common four-step pattern:

1. Analysis – Extract new information about the domain structure.
2. Generation – Build macro-operators based on the previously acquired information.
3. Filtering – Select the most promising macro-operators.
4. Planning – Use the selected macro-operators to improve planning in future problems.

### 1.1 Component Abstraction – Enhanced Domain

The first method produces a small set of macro-operators from the PDDL formulations of a domain and several training problems. The macro-operators are added to the initial domain formulation, resulting in an enhanced domain expressed in the same description language. The definitions of the enhanced domain and new problem instances can be given as input to any planner, with no need to implement additional support for macro-operators (Botea, Müller, & Schaeffer, 2004b). We call this approach *CA-ED* for *Component Abstraction – Enhanced Domain*.

Figure 1 shows the general architecture of CA-ED. The box Abstraction in the figure includes steps 1 – 3 above. Step 1 uses *component abstraction*, a technique that exploits permanent relationships between low-level features of a problem. Low-level features (i.e., constant symbols) linked by static facts (i.e., facts that remain true during planning) form a more complex unit called an abstract component.

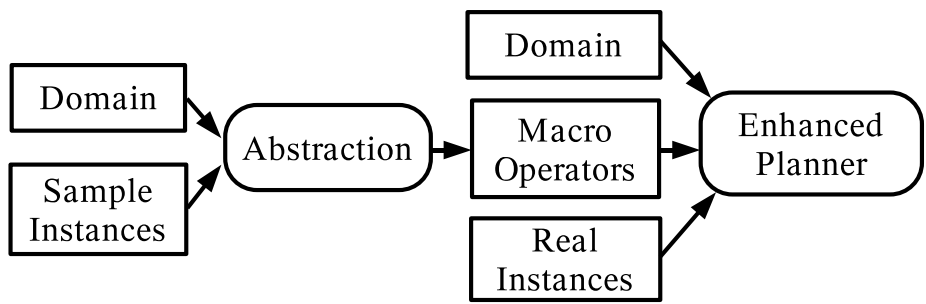

Figure 2: The general architecture of SOL-EP. Enhanced Planner means a planner with capabilities to handle macros.

At step 2, local analysis of abstract components builds macros that can speed up planning. CA-ED generates macros using a forward search process in the space of macro operators. A macro operator is built as an ordered sequence of operators linked through a mapping of the operators' variables. Applying a macro operator is semantically equivalent to applying all contained operators in the given order, respecting the macro's variable mapping and the interactions between preconditions and effects.

At step 3 (filtering), a set of heuristic rules is used to prune the search space and generate only macros that are likely to be useful. Macros are further filtered dynamically, based on their performance in solving training problems, and only the most effective ones are kept for future use.

The best macro operators that this method generates are added as new operators to the initial PDDL domain formulation, enhancing the initial set of operators. Hence, we need *complete* macro-operator definitions, including precondition and effect formulas. Expressing these formulas starting from the contained operators is easy in STRIPS, but hard in larger PDDL subsets such as ADL, where the preconditions and the effects of the contained operators can interact in complex ways. See Section 3.1 for a detailed explanation. In CA-ED no work is required to implement step 4, since the planner makes no distinction between a macro operator and a normal operator. Once the enhanced domain formulation is available in standard STRIPS, any planner can be used to solve problem instances.

The architecture of CA-ED has two main limitations. First, component abstraction can currently be applied only to domains with static facts in their formulation. Second, adding macros to the original domain definition is limited to STRIPS domains.

### 1.2 Solution – Enhanced Planner

The second abstraction method presented in this article does not suffer from the above limitations, and is applicable to a larger class of problems. We call this approach *SOL-EP*, which stands for *Solution – Enhanced Planner*. SOL-EP extracts macros from solutions of training problems and uses them in a planner enhanced with capabilities to handle macros. The general architecture of this approach is shown in Figure 2. As before, the module Abstraction implements steps 1 – 3. Instead of using static facts and component abstraction as in CA-ED, step 1 in SOL-EP processes the solutions of training problems. To extend applicability from STRIPS to ADL domains, a different macro representation is used as compared to CA-ED. A SOL-EP macro is represented as a sequence of operators and a mapping of the operators' variables rather than a compilation into a single operator with complete definition of its precondition and effects. As shown in Section 3.1, for ADL domains, it is impractical to use macros with complete definition.

For this reason, SOL-EP macros cannot be added to the original domain formulation as new operators anymore. They are distinct input data for the planner, and for step 4 the planner is enhanced with code to handle macro operators. Since SOL-EP is more general, we used this approach in the fourth planning competition IPC-4.

We implemented the ideas presented in this article in MACRO-FF, an adaptive planning system developed on top of FF version 2.3 (Hoffmann & Nebel, 2001). FF 2.3 is a state-of-the-art fully automatic planner that uses a heuristic search approach. The solving mechanism of FF has two main phases: *preprocessing* and *search*. The preprocessing phase builds data structures needed at search time. All operators are instantiated into ground actions, and all predicates are instantiated into facts. For each action, pointers are stored to all precondition facts, all add-effect facts, and all delete-effect facts. Similarly, for each fact $f$, pointers are stored to all actions where $f$ is a precondition, to all actions where $f$ is an add effect, and to all actions where $f$ is a delete effect. This information is instantly available at run-time, when states are evaluated with the relaxed graphplan heuristic.

MACRO-FF adds the ability to automatically learn and use macro-actions, with the goal of improving search. MACRO-FF also includes engineering enhancements that can reduce space and CPU time requirements that were performance bottlenecks in some of the test problems. The engineering enhancements affect neither the number of expanded nodes nor the quality of found plans.

The contributions of this article include a detailed presentation of MACRO-FF. We present and compare two methods that automatically create and use macro-operators in domain-independent AI planning. Experimental evaluation is focused on several main directions. First, the impact of the engineering enhancements is analyzed. Then we evaluate how SOL-EP macros implemented in the competition system can improve planning. These experiments use as testbeds domains used in IPC-4. Finally, we compare the two abstraction methods on test instances where both techniques are applicable, and evaluate them against planning with no macros.

The rest of the paper is structured as follows: The next two sections describe CA-ED and SOL-EP respectively. Section 4 summarizes the implementation enhancements that we added to FF. We present experimental results and evaluate our methods in Section 5. In Section 6 we briefly review related work and discuss the similarities and differences with our work. The last section contains conclusions and ideas for future work.

## 2. Enhancing a Domain with Macros based on Component Abstraction

The first part of this section introduces component abstraction. The topic of the second part is CA-ED macro-operators.

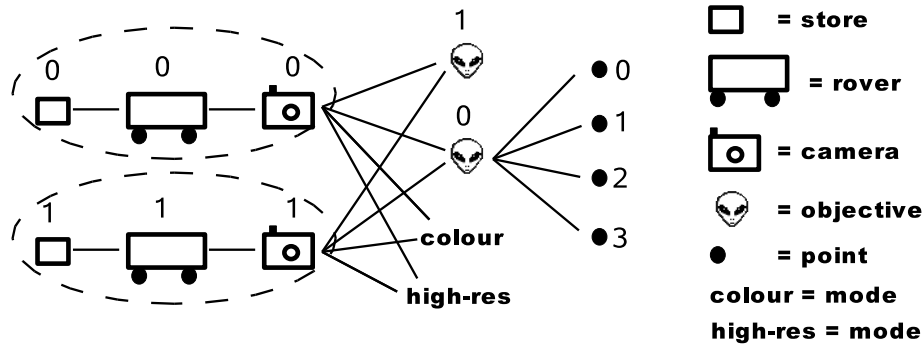

Figure 3: Static graph of a Rovers problem.

### 2.1 Component Abstraction

Component abstraction is a technique that groups related low-level constants of a planning problem into more abstract entities called *abstract components* or, shorter, components. The idea is similar to how humans can group features connected through static relationships into one more abstract unit. For example, a robot that carries a hammer could be considered a single component, which has mobility as well as maintenance skills. Such a component can become a permanent object in the representation of the world, provided that no action can invalidate the static relation between the robot and the hammer.

Component abstraction is a two-step procedure:

1. Build the problem's *static graph*, which models permanent relationships between constant symbols of a problem.

2. Build abstract components with a clustering procedure. Formally, an abstract component is a connected subgraph of the static graph.

#### 2.1.1 BUILDING THE STATIC GRAPH OF A PROBLEM

A static graph models static relationships between constant symbols of a problem. Nodes are constant symbols, and edges correspond to *static facts* in the problem definition. Following standard terminology, a fact is an instantiation of a domain predicate, i.e., a predicate whose parameters have all been instantiated with constant symbols. A fact $f$ is static for a problem $p$ if $f$ is part of the initial state of $p$ and no operator can delete it.

Each constant that is an argument of at least one static fact defines a node in the static graph. All constants in a fact are linked pairwise. All edges in the graph are labeled with the name of the corresponding predicate.

We use a Rovers problem as an example of how component abstraction works. In this domain, *rovers* can be equipped with *cameras* and *stores* where rock and soil samples can be collected and analyzed. Rovers have to gather pictures and data about rock and soil samples, and report them to their base. For more information about the Rovers domain, see the work of Long and Fox (2003). Figure 3 shows the static graph of the sample problem. The nodes include two stores (STORE0 and STORE1), two rovers (ROVER0 and ROVER1), two photo cameras (CAM0 and CAM1), two objectives (OBJ0 and OBJ1), two camera modes (COLOUR and HIGH-RES), and four waypoints (POINT0,... POINT3). The edges correspond to the static predicates (STORE-OF ?S - STORE ?R - ROVER), (ON-BOARD ?C - CAMERA ?R - ROVER), (SUPPORTS ?C - CAMERA ?M - MODE), (CALIBRATION-TARGET ?C - CAMERA ?O - OBJECTIVE), and (VISIBLE-FROM ?O - OBJECTIVE ?W - WAYPOINT).

The two marked clusters in the left are examples of abstract components generated by this method. Each component is a rover equipped with a camera and a store. A more detailed and formal explanation is provided in the following paragraphs.

To identify static facts necessary to build the static graph, the set of domain operators $\mathcal{O}$ is used to partition the predicate set $\mathcal{P}$ into two disjoint sets, $\mathcal{P} = \mathcal{P}_F \cup \mathcal{P}_S$, corresponding to *fluent* and *static* predicates. An operator $o \in \mathcal{O}$ is represented as a structure

$$o = (V(o), P(o), A(o), D(o)),$$

where $V(o)$ is the variable set, $P(o)$ is the precondition set, $A(o)$ is the set of add effects, and $D(o)$ is the set of delete effects. A predicate $p$ is fluent if $p$ is part of an operator's effects (either positive or negative):

$$p \in \mathcal{P}_F \Leftrightarrow \exists o \in \mathcal{O} : p \in A(o) \cup D(o).$$

Otherwise, $p$ is static, denoted by $p \in \mathcal{P}_S$.

In a domain with hierarchical types, instances of the same predicate can be both static and fluent. Consider the Depots domain, a combination of Logistics and Blocksworld, which was used in the third international planning competition (Long & Fox, 2003). This domain has such a type hierarchy. Type LOCATABLE has four atomic sub-types: PALLET, HOIST, TRUCK, and CRATE. Type PLACE has two atomic sub-types: DEPOT and DISTRIBUTOR. Predicate (AT ?L - LOCATABLE ?P - PLACE), which indicates that object ?L is located at place ?P, corresponds to eight specialized predicates at the atomic type level. Predicate (AT ?P - PALLET ?D - DEPOT) is static, since there is no operator that adds, deletes, or moves a pallet. However, predicate (AT ?C - CRATE ?D - DEPOT) is fluent. For instance, the LIFT operator deletes a fact of this type.

To address the issue of hierarchical types, we use a domain formulation where all types are expressed at the lowest level in the hierarchy. We expand each predicate into a set of *low-level predicates* whose arguments have low-level types. Similarly, *low-level operators* have variable types from the lowest hierarchy level. Component abstraction and macro generation are done at the lowest level. After building the macros, we restore the type hierarchy of the domain. When possible, we replace a set of two or more macro operators that have low-level types with one equivalent macro operator with hierarchical types. In this way, macros respect the same definition style (with respect to hierarchical types) as the rest of the domain operators. For planners that pre-instantiate all operators, such as FF, the existence of hierarchical types is not relevant. Before searching for a solution, all operators are instantiated into ground actions whose arguments have low-level types.

Facts corresponding to static predicates are called static facts. In our current implementation we ignore static predicates that are unary [1] or contain two or more variables of the same type. The latter kind of facts are often used to model topological relationships, and can lead to large components.

---

1. In fact, in many current domains, unary static facts have been replaced by types associated with variables.

| Step | Current Predicate | Used. Pred. | COMPONENT0 | | COMPONENT1 | |
|---|---|---|---|---|---|---|
| | | | Consts | Facts | Consts | Facts |
| 1 | | | CAM0 | | CAM1 | |
| 2 | (SUPPORTS ?C - CAMERA ?M - MODE) | NO | CAM0 | | CAM1 | |
| 3 | (CALIBR-TARGET ?C - CAMERA ?O - OBJECTIVE) | NO | CAM0 | | CAM1 | |
| 4 | (ON-BOARD ?C - CAMERA ?R - ROVER) | YES | CAM0 ROVER0 | (ON-BOARD CAM0 ROVER0) | CAM1 ROVER1 | (ON-BOARD CAM1 ROVER1) |
| 5 | (STORE-OF ?S - STORE ?R - ROVER) | YES | CAM0 ROVER0 STORE0 | (ON-BOARD CAM0 ROVER0) (STORE-OF STORE0 ROVER0) | CAM1 ROVER1 STORE1 | (ON-BOARD CAM1 ROVER1) (STORE-OF STORE1 ROVER1) |

Table 1: Building abstract components for the Rovers example.

### 2.1.2 BUILDING ABSTRACT COMPONENTS

Abstract components are built as connected subgraphs of the static graph of a problem. Clustering starts with abstract components of size 1, containing one node each, that are generated based on a domain type $t$, called the *seed* type. For each node with type $t$ in the static graph, a new abstract component is created. Abstract components are then iteratively extended with a greedy approach.

Next we detail how the clustering procedure works on the example, and then provide a more formal description, including pseudo-code. As said before, Figure 3 shows the two abstract components built by this procedure. The steps of the clustering are summarized in Table 1, and correspond to the following actions:

1. Choose a seed type (CAMERA in this example), and create one abstract component for each constant of type CAMERA: COMPONENT0 contains CAM0, and COMPONENT1 contains CAM1. Next, iteratively extend the components created at this step. One extension step uses a static predicate that has at least one variable type already encoded into the components.

2. Choose the predicate (SUPPORTS ?C - CAMERA ?M - MODE), which has a variable of type CAMERA. To avoid ending up with one large component containing the whole graph, merging two existing components is not allowed. Hence a check is performed whether the static facts based on this predicate keep the existing components separated. These static facts are (SUPPORTS CAM0 COLOUR), (SUPPORTS CAM0 HIGH-RES), (SUPPORTS CAM1 COLOUR), and (SUPPORTS CAM1 HIGH-RES). The test fails, since constants COLOUR and HIGH-RES would be part of both components. Therefore, this predicate is not used for component extension (see the third column of Table 1).

3. Similarly, the predicate (CALIBRATION-TARGET ?C - CAMERA ?O - OBJECTIVE), which would add the constant OBJ1 to both components, is not used for extension.

4. The predicate (ON-BOARD ?C - CAMERA ?R - ROVER) is tried. It merges no components, so it is used for component extension. The components are expanded as shown in Table 1, Step 4.

5. The predicate (STORE-OF ?S - STORE ?R - ROVER), whose type ROVER has previously been encoded into the components, is considered. This predicate extends the components as presented in Table 1, Step 5.

After Step 5 is completed, no further component extension can be performed. There are no other static predicates using at least one of the component types to be tried for further extension. At this moment the quality of the decomposition is evaluated. In this example it is satisfactory (see discussion below), and the process terminates. Otherwise, the decomposition process restarts with another domain type.

The quality of a decomposition is evaluated according to the size of the built components, where size is defined as the number of low-level types in a component. In our experiments, we limited size to values between 2 and 4. The lower limit is trivial, since an abstract component should combine at least two low-level types. The upper limit was set heuristically, to prevent the abstraction from building just one large component. These relatively small values are also consistent with the goal of limiting the size and number of macro operators. We discuss this issue in more detail in Section 2.2.

Figure 4 shows pseudo-code for component abstraction. $Types(g)$ contains all types of the constant symbols used as nodes in $g$. Given a type $t$, $Preds(t)$ is the set of all static predicates that have a parameter of type $t$. Given a static predicate $p$, $Types(p)$ includes the types of its parameters. $Facts(p)$ are all facts instantiated from $p$.

Each iteration of the main loop tries to build components starting from a seed type $t \in Types(g)$. The sets $Open$, $Closed$, $Tried$, and $AC$ are initialized to $\emptyset$. Each graph node of type $t$ becomes the seed of an abstract component (method *createComponent*). The components are greedily extended by adding new facts and constants, such that no constant is part of any two distinct components. The method $predConnectsComponents(p, AC)$ verifies if any fact $f \in Facts(p)$ merges two distinct abstract components in $AC$.

Method $extendComponents(p, AC)$ extends the existing components using all static facts $f \in Facts(p)$. For simplicity, assume that a fact $f$ is binary and has constants $c_1$ and $c_2$ as arguments. Given a component $ac$, let $Nodes(ac)$ be its set of constants (subgraph nodes) and $Facts(ac)$ its set of static facts (subgraph edges). In the most general case, four possible relationships can exist between the abstract components and elements $f$, $c_1$, and $c_2$:

1. Both $c_1$ and $c_2$ already belong to the same abstract component $ac$:

$$\exists(ac \in AC) : c_1 \in Nodes(ac) \land c_2 \in Nodes(ac).$$

   In this case, $f$ is added to $ac$ as a new edge.

2. Constant $c_1$ is already part of an abstract component $ac$ (i.e., $c_1 \in Nodes(ac)$) and $c_2$ is not assigned to a component yet. Now $ac$ is extended with $c_2$ as a new node and $f$ as a new edge between $c_1$ and $c_2$.

3. If neither $c_1$ nor $c_2$ are part of a previously built component, a new component containing $f$, $c_1$ and $c_2$ is created and added to $AC$.

MACRO-FF: IMPROVING AI PLANNING WITH AUTOMATICALLY LEARNED MACRO-OPERATORS

```
componentAbstraction(Graph g) {
  for (each t ∈ Types(g) chosen in random order) {
    resetAllStructures();
    Open ← t;
    for (each c_i ∈ Nodes(g) with type t)
      AC ← createComponent(c_i);
    while (Open ≠ ∅) {
      t_1 ← Open;
      Closed ← t_1;
      for (each p ∈ Preds(t_1) \ Tried)
        Tried ← p;
        if ¬(predConnectsComponents(p, AC)) {
          extendComponents(p, AC);
          for (each t_2 ∈ Types(p))
            if (t_2 ∉ Open ∪ Closed)
              Open ← t_2;
        }
    }
    if (evaluateDecomposition() = OK)
      return AC;
  }
  return ∅;
}
```

Figure 4: Component abstraction in pseudo-code.

4. Constants $c_1$ and $c_2$ belong to two distinct abstract components:

$$\exists (ac_1, ac_2) : c_1 \in Nodes(ac_1) \land c_2 \in Nodes(ac_2) \land ac_1 \neq ac_2.$$

While possible in general, this last alternative never occurs at the point where the method *extendComponents* is called. This is ensured by the previous test with the method *predConnectsComponents*.

Consider the case when a static graph has two disconnected (i.e., with no edge between them) subgraphs $sg_1$ and $sg_2$ such that $Types(sg_1) \cap Types(sg_2) = \emptyset$. In such a case, the algorithm shown in Figure 4 finds abstract components only in the subgraph that contains the seed type. To perform clustering on the whole graph, the algorithm has to be run on each subgraph separately.

Following the standard of typed planning domains, abstract components are assigned *abstract types*. Figure 5 shows the abstract type assigned to the components of our example. As shown in this figure, the abstract type of an abstract component is a graph obtained from the component graph by changing the node labels. The constant symbols used as node labels have been replaced with their low-level types (e.g., constant CAM0 has been replaced by its type CAMERA).

589589

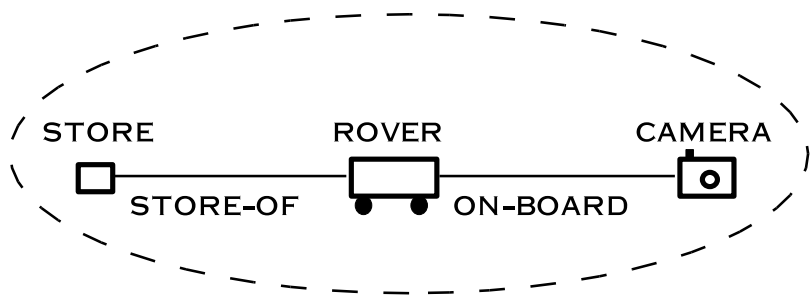

Figure 5: Abstract type in Rovers.

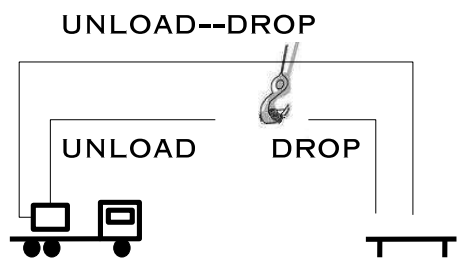

Figure 6: Example of a macro in Depots.

Our example also shows that components with identical structure have the same abstract type. Identical structure is a strong form of graph isomorphism, which preserves the edge labels as well as the types of constants used as node labels. A fact $f = f(c^1, ..., c^k) \in Facts(ac)$ is a predicate whose variables have been instantiated to constants $c^i \in Nodes(ac)$.

Two abstract components $ac_1$ and $ac_2$ have identical structure if:

1. $|Nodes(ac_1)| = |Nodes(ac_2)|$; and

2. $|Facts(ac_1)| = |Facts(ac_2)|$; and

3. there is a bijective mapping $p : Nodes(ac_1) \to Nodes(ac_2)$ such that

   - $\forall c \in Nodes(ac_1) : Type(c) = Type(p(c))$;
   - $\forall f(c_1^1, ..., c_1^k) \in Facts(ac_1) : f(p(c_1^1), ..., p(c_1^k)) \in Facts(ac_2)$;
   - $\forall f(c_2^1, ..., c_2^k) \in Facts(ac_2) : f(p^{-1}(c_2^1), ..., p^{-1}(c_2^k)) \in Facts(ac_1)$;

### 2.2 Creating Macro-Operators

A macro-operator $m$ in CA-ED is formally equivalent to a normal operator: it has a set of variables $V(m)$, a set of preconditions $P(m)$, a set of add effects $A(m)$, and a set of delete effects $D(m)$. Figure 6 shows an example of a macro in Depots. Figure 7 shows complete STRIPS definitions for this macro and the operators that it contains.

Macro operators are obtained in two steps, which are presented in detail in the remaining part of this section. First, an extended set of macros is built and next the macros are

```
 (:action UNLOAD—DROP
   :parameters
     (?h - hoist ?c - crate ?t - truck ?p - place ?s - surface)
   :precondition
     (and    (at ?h ?p) (in ?c ?t) (available ?h)
             (at ?t ?p) (clear ?s) (at ?s ?p))
   :effect
     (and    (not (in ?c ?t)) (not (clear ?s))
             (at ?c ?p) (clear ?c) (on ?c ?s))
)
(:action UNLOAD
   :parameters
     (?x - hoist ?y - crate ?t - truck ?p - place)
   :precondition
     (and    (in ?y ?t) (available ?x) (at ?t ?p) (at ?x ?p))
   :effect
     (and    (not (in ?y ?t)) (not (available ?x)) (lifting ?x ?y))
)
(:action DROP
   :parameters
     (?x - hoist ?y - crate ?s - surface ?p - place)
   :precondition
     (and    (lifting ?x ?y) (clear ?s) (at ?s ?p) (at ?x ?p))
   :effect
     (and    (available ?x) (not (lifting ?x ?y)) (at ?y ?p)
             (not (clear ?s)) (clear ?y) (on ?y ?s))
)
```

Figure 7: STRIPS definitions of macro UNLOAD—DROP and the operators that it contains.

filtered in a quick training process. Since empirical evidence indicates that the extra information added to a domain definition should be quite small, the methods described next tend to minimize the number of macros and their size, measured by the number of variables, preconditions and effects. Static macro generation uses many constraints for pruning the space of macro operators, and discards large macros. Finally, dynamic filtering keeps only a few top performing macros for solving future problems.

### 2.2.1 MACRO GENERATION

For an abstract type *at*, macros are generated by performing a forward search in the space of macro operators. Macros perform local processing within a component of type *at*, according to the locality rule detailed below.

The root state of the search represents an empty macro (i.e., empty sets of operators, variables, preconditions, and effects). Each search step appends an operator to the current

```
void addOperatorToMacro(operator o, macro m, variable-mapping vm) {
  for (each precondition p ∈ P(o)) {
    if (p ∉ A(m) ∪ P(m))
      P(m) = P(m) ∪ {p};
  }
  for (each delete effect d ∈ D(o)) {
    if (d ∈ A(m))
      A(m) = A(m) − {d};
    D(m) = D(m) ∪ {d};
  }
  for (each add effect a ∈ A(o)) {
    if (a ∈ D(m))
      D(m) = D(m) − {a};
    A(m) = A(m) ∪ {a};
  }
}
```

Figure 8: Adding operators to a macro.

macro, and fixes the variable mapping between the new operator and the macro. Adding a new operator $o$ to a macro $m$ modifies $P(m)$, $A(m)$, and $D(m)$ as shown in Figure 8. Even if not explicitely shown in the figure, the variable mapping $vm$ in the procedure is used to check the identity between operator's predicates and macro's predicates (e.g., in $p \notin A(m) \cup P(m)$). Two predicates are considered identical if they have the same name and the same set of parameters. The variable mapping $vm$ tells what variables (parameters) are common in both the macro and the new operator.

The search is selective: it includes a set of rules for pruning the search tree and for validating a built macro operator. Validated macros are goal states in this search space. The search enumerates all valid macro operators. The following pruning rules are used for static filtering:

- The *negated precondition rule* prunes operators with a precondition that matches one of the current delete effects of the macro operator. This rule avoids building incorrect macros where a predicate should be both true and false.

- The *repetition rule* prunes operators that generate cycles. A macro containing a cycle is either useless, producing an empty effect set, or it can be written in a shorter form by eliminating the cycle. A cycle in a macro is detected when the effects of the first $k_1$ operators are the same as for the first $k_2$ operators, with $k_1 < k_2$. In particular, if $k_1 = 0$ then the first $k_2$ operators have no effect.

- The *chaining rule* requires that for consecutive operators $o_1$ and $o_2$ in a macro, the preconditions of $o_2$ must include at least one positive effect of $o_1$. This rule is motivated by the idea that the action sequence of a macro should have a coherent meaning.

```
(:action TAKE-IMAGE
  :parameters
    (?r - rover ?p - waypoint ?o - objective ?i - camera ?m - mode)
  :precondition
    (and   (calibrated ?i ?r) (on-board ?i ?r) (equipped-for-imaging ?r)
           (supports ?i ?m) (visible-from ?o ?p) (at ?r ?p))
  :effect
    (and   (have-image ?r ?o ?m) (not (calibrated ?i ?r)))
)
(:action TAKE-IMAGE—TAKE-IMAGE
  :parameters
    (?r0 - rover ?p - waypoint ?o - objective ?i0 - camera ?m - mode
    ?r1 - rover ?i1 - camera)
  :precondition
    (and   (calibrated ?i0 ?r0) (on-board ?i0 ?r0) (equipped-for-imaging ?r0)
           (calibrated ?i1 ?r1) (on-board ?i1 ?r1) (equipped-for-imaging ?r1)
           (supports ?i0 ?m) (visible-from ?o ?p) (at ?r0 ?p)
           (supports ?i1 ?m) (at ?r1 ?p))
  :effect
    (and   (have-image ?r0 ?o ?m) (not (calibrated ?i0 ?r0))
           (have-image ?r1 ?o ?m) (not (calibrated ?i1 ?r1)))
)
```

Figure 9: Operator TAKE-IMAGE and macro-operator TAKE-IMAGE—TAKE-IMAGE in Rovers. This macro is rejected by the locality rule.

- We limit the size of a macro by imposing a maximal length and a maximal number of preconditions. Similar constraints could be added for the number of variables or effects, but we found this unnecessary. Limiting the number of preconditions indirectly limits the number of variables and effects. Large macros are generally undesirable, as they can significantly increase the preprocessing costs and the cost per node in the planner's search.

- The *locality rule* is meant to prune macros that change two or more abstract components at the same time. All *local static preconditions* of an acceptable macro are part of the same abstract component. Given an abstract type $at$ and a macro $m$, let the *local static preconditions* be the static predicates that are part of both $m$'s preconditions and $at$'s edges. Local static preconditions and their parameters in $m$'s definition define a graph structure (different variable bindings for the operators that compose $m$ can create different graph structures). To implement the idea of locality we require that this graph is isomorphic with a subgraph of $at$.

As an example of the locality rule, consider the Rovers abstract type $at$ in Figure 5 and the macro $m$ TAKE-IMAGE—TAKE-IMAGE shown in Figure 9 (this figure also shows the definition of the TAKE-IMAGE operator). Intuitively, $m$ involves two components, since two distinct cameras and two distinct rovers are part of the macro's variables. We show that this macro is rejected by the locality rule. The graph corresponding to the local static preconditions of $m$ and $at$ is shown in Figure 10. Obviously, this is not a subgraph of $at$'s graph shown in Figure 5, so $m$ is rejected.

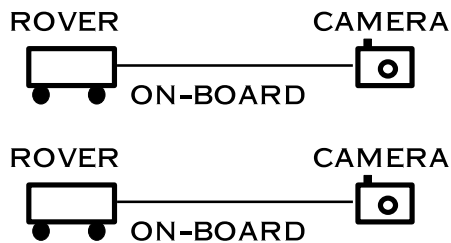

Figure 10: Local static preconditions of macro TAKE-IMAGE—TAKE-IMAGE with respect to the abstract type in Figure 5. As the picture shows, these correspond to a graph with 4 nodes and 2 edges.

### 2.2.2 Macro Ranking and Filtering

The goal of ranking and filtering is to reduce the number of macros and use only the most efficient ones for solving problems. The overhead of poor macros can outweight their benefit. This is known under the name of the *utility problem* (Minton, 1988). In CA-ED, adding more operators to a domain increases the *preprocessing* costs and the cost per node in the planner's search.

We used a simple but efficient and practical approach to dynamic macro filtering to select a small set of macro operators. We count how often a macro operator is instantiated as an action in the problem solutions found by the planner. The more often a macro has been used in the past, the greater the chance that the macro will be useful in the future.

For ranking, each macro operator is assigned a weight that estimates its efficiency. All weights are initialized to 0. Each time a macro is present in a plan, its weight is increased by the number of occurrences of the macro in the plan (*occurrence points*), plus 10 *bonus points*. No effort was spent on tuning parameters such as the bonus. For *common macros* that are part of *all* solutions of training problems, any bonus value $v \geq 0$ will produce the same ranking among these common macros. No matter what the value $v$ is, each common macro will receive $v \times T$ bonus points, where $T$ is the number of training problems. Hence the occurrence points decide the relative ranking of common macros.

We use the simplest problems in a domain for training. For these simple problems, we use all macro operators, giving each macro a chance to participate in a solution plan and increase its weight. After the training phase, the best macro operators are selected to become part of the enhanced domain definition. In experiments, 2 macros, each containing two steps, were added as new operators to the initial sets of 9 operators in Rovers, and 5 operators in Depots and Satellite. In these domains, such a small amount of extra-information was observed to be a good tradeoff between the benefits and the additional pre-processing and run-time costs. In more difficult domains, possibly with larger initial sets of operators, using more macros would probably be beneficial.

## 3. Using Macros from Solutions in an Enhanced Planner

In this section we introduce SOL-EP, the macro system that we used in the fourth international planning competition. We start with a motivation in Section 3.1, and then describe the method in the following two sections. SOL-EP follows the same four-step pattern as before, but can be applied to more general classes of problems. Section 3.2 describes steps 1 – 3, and Section 3.3 shows step 4. Section 3.4 concludes this section with a discussion.

### 3.1 Motivation

SOL-EP was designed with the goal of eliminating the main limitations of CA-ED. Specifically, we wanted to extend the applicability of CA-ED to larger classes of domains. Since CA-ED generates macros based on component abstraction, its applicability is limited to domains with static predicates in their definition. SOL-EP generates macros from solutions of sample problems, with no restrictions caused by the nature of a domain's predicates.

Furthermore, CA-ED is limited to relatively simple subsets of PDDL such as STRIPS. Since CA-ED adds macros as new operators to the original domain, complete definitions of macros, including precondition and effect formulas, are required. These formulas are easy to obtain in STRIPS, as shown in Figures 7 and 8. However, adding macros to an ADL domain file becomes unfeasible in practice for two main reasons. First, the precondition and effect formulas of a macro are hard to infer from the formulas of contained operators. Second, even if the previous issue is solved and a macro with complete definition is added to a domain, the costs for pre-instantiating it into ground macro-actions can be large.

To illustrate how challenging the formula inference is in ADL, consider the example in Figure 11, which shows operator MOVE from the ADL Airport domain used in IPC-4. The preconditions and the effects of this operator are quite complex formulas that include quantifiers, implications and conditional effects. Assume we want to compose a macro that applies two MOVE actions in a row with a given parameter mapping. To achieve a complete definition of macro MOVE MOVE, its precondition and effect formulas would have to be automatically composed by analyzing how the preconditions and effects of the two contained operators interact. We could not find a straight-forward way to generate a macro's formulas, so we decided to move towards an alternative solution that is presented later in this subsection.

Even if the above issue is solved and macros can be added as new domain operators, pre-instantiating a macro into ground actions can be costly. Many top-level planners, including FF, pre-instantiate the domain operators into all possible ground actions that might be applied in the problem instance at hand. The cost of instantiating one operator is exponential in the total number of parameters and quantifier variables. Macros tend to have larger numbers of parameters and quantifiers and therefore their instantiation can significantly increase the total preprocessing costs. ADL Airport is a good illustration of how important this effect can be. As shown in Section 5.2, the preprocessing is so costly as compared to the

```
(:action move
 :parameters
   (?a - airplane ?t - airplanetype ?d1 - direction ?s1 ?s2 - segment ?d2 - direction)
 :precondition
   (and   (has-type ?a ?t) (is-moving ?a)
          (not (= ?s1 ?s2))
          (facing ?a ?d1) (can-move ?s1 ?s2 ?d1)
          (move-dir ?s1 ?s2 ?d2) (at-segment ?a ?s1)
          (not
             (exists    (?a1 - airplane)
                        (and (not (= ?a1 ?a)) (blocked ?s2 ?a1))))
          (forall   (?s - segment)
                        (imply   (and   (is-blocked ?s ?t ?s2 ?d2)
                                        (not (= ?s ?s1)))
                                 (not (occupied ?s))))
   )
 :effect
   (and    (occupied ?s2) (blocked ?s2 ?a)
           (not (occupied ?s1))
           (when   (not (is-blocked ?s1 ?t ?s2 ?d2))
                        (not (blocked ?s1 ?a)))
           (when   (not (= ?d1 ?d2))
                        (not (facing ?a ?d1)))
           (not (at-segment ?a ?s1))
           (forall   (?s - segment)
                        (when   (is-blocked ?s ?t ?s2 ?d2)
                                 (blocked ?s ?a)))
           (forall   (?s - segment)
                        (when   (and   (is-blocked ?s ?t ?s1 ?d1)
                                        (not (= ?s ?s2))
                                        (not (is-blocked ?s ?t ?s2 ?d2)))
                                 (not (blocked ?s ?a))))
           (at-segment ?a ?s2)
           (when (not (= ?d1 ?d2))
           (facing ?a ?d2))
   )
)
```

Figure 11: Operator MOVE in ADL Airport.

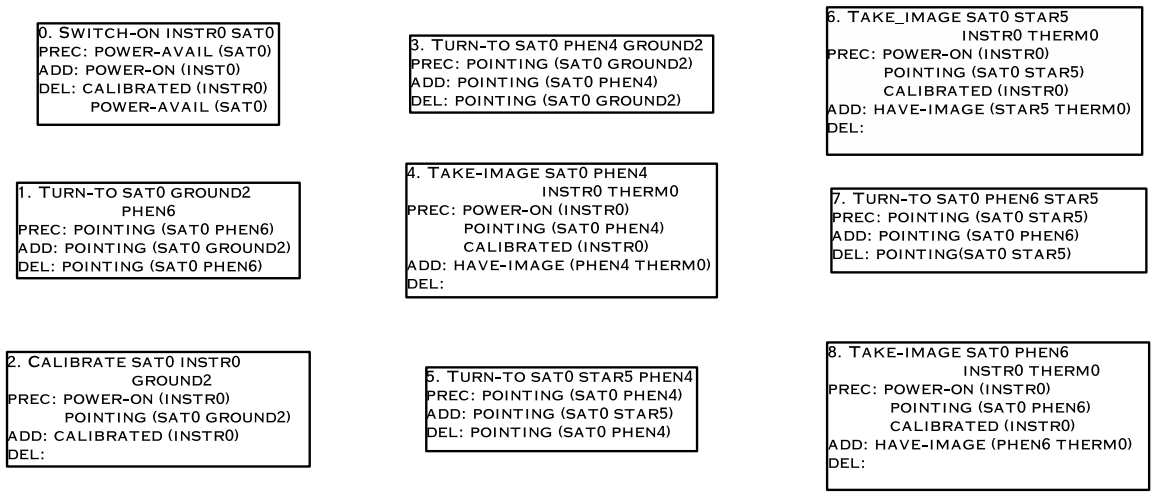

Figure 12: The solution steps of problem 1 in the Satellite benchmark.

main search that it dominates the total cost of solving a problem in this domain. Further increasing the preprocessing costs with new operators is not desirable in such domains.

Our solution for ADL macros is to represent a SOL-EP macro as a list of atomic actions. Precondition and effect formulas are not explicitly provided. Rather, they are determined at run-time, when a macro is dynamically instantiated by applying its action sequence.

The benchmarks used in IPC-4 emphasize the need to address the issues described above. Many competition domains were provided in both STRIPS and ADL formulations. The "main" definition was in ADL and, for planners that could not take ADL domains as input, STRIPS compilations of each ADL domain were provided. We could only run our system on ADL domains. The reason is that in STRIPS compilations of ADL domains, a distinct domain file was generated for each problem instance. However, our learning approach requires several training problems for each domain.

### 3.2 Generating Macros

As a running example, we will use the solution plan for problem 1 in the Satellite domain shown in Figure 12. For each step, the figure shows the order in the linear plan, the action name, the argument list, the preconditions, and the effects. To keep the picture simple, we ignore static preconditions of actions. Static facts never occur as action effects, and therefore do not affect the interactions between preconditions and effects of actions.

In SOL-EP, macro-operators are extracted from the solutions of the training problems. Each training problem is first solved with no macros in use. The found plan can be represented as a *solution graph*, where each node represents a plan step (action), and edges model interactions between solution steps. Building the solution graph is step 1 (analysis) in our general four-step pattern. In IPC-4 we used a first implementation of the solution graph, that considers interactions only between two consecutive actions of a plan. Here an interaction is defined if the two actions have at least one common argument, or at least one action has no arguments at all. Hence the implementation described in this article extracts only such two-action sequences as possible macros.

The macro-actions extracted from a solution are translated into macro-operators by replacing their instantiated arguments with generic variables. This operation preserves the relative mapping between the arguments of the contained actions. Macro-actions with different sets of arguments can result in the same macro-operator. For the Satellite solution in Figure 12, the sequence TURN-TO followed by TAKE-IMAGE occurs three times. After replacing the constant arguments with generic variables, all occurrences yield the same macro-operator.

There are many pairs of actions in a solution, and a decision must be made as to which ones are going to beneficial as macro-operators in a search. Macros are statically filtered according to the rules of Section 2.2.1 excluding the limitation of the number of preconditions, which is not critical in this algorithm, and the locality rule. Also, as said before, we use a different version of the chaining rule. We request that the operators of a macro have common variables, unless an operator has 0 parameters.

Macro-operators are stored in a global list ordered by their *weight*, with smaller being better. Weights are initialized to 1.0 and updated in a dynamic ranking process using a gradient-descent method.

For each macro-operator $m$ extracted from the solution of a training problem, we re-solve the problem with $m$ in use. Let $L$ be the solution length when no macros are used, $N$ the number of nodes expanded to solve the problem with no macros, and $N_m$ the number of expanded nodes when macro $m$ is used. Then we use the difference $N - N_m$ to update $w_m$, the weight of macro $m$. Since $N - N_m$ can take arbitrarily large values, we map it to a new value in the interval $(-1, 1)$ by

$$\delta_m = \sigma(\frac{N - N_m}{N})$$

where $\sigma$ is the sigmoid function

$$\sigma(x) = \frac{2}{1 + e^{-x}} - 1.$$

Function $\sigma$ generates the curve shown in Figure 13. This particular definition of $\sigma$ was chosen because it is symmetric in $(0,0)$ (i.e., $\sigma(x) = -\sigma(-x)$) and bounded within the interval $(-1, 1)$. In particular, the symmetry property ensures that, if $N_m = N$, than the weight update of $m$ at the current training step is 0. The size of the boundary interval has no effect on the ranking procedure, it only scales all weight updates by a constant multiplicative factor. We used a sigmoid function bounded to $(-1, 1)$ as a canonical representation, which limits the absolute value of $\delta_m$ between 0 and 1.

The update formula also contains a factor that measures the difficulty of the training instance. The harder the problem, the larger the weight update should be. We use as the difficulty factor the solution length $L$ rather than $N$, since the former has a smaller variance over a training problem set. The formula for updating $w_m$ is

$$w_m = w_m - \alpha \delta_m L$$

where $\alpha$ is a small constant (0.001 in our implementation). The value of $\alpha$ does not affect the ranking of macros. It was used only to keep macro weights within the vicinity of 1. See the second part of Section 3.4 for a comparison between CA-ED's frequency-based ranking and SOL-EP's gradient-descent ranking.

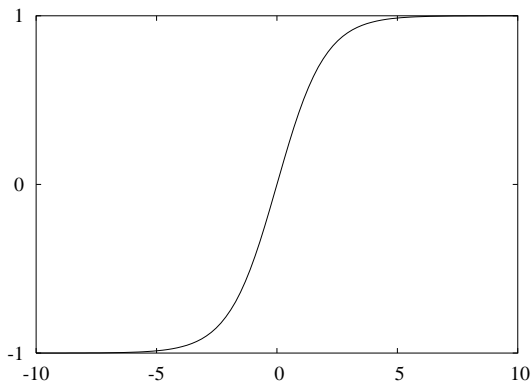

Figure 13: Sigmoid function.

In CA-ED, only two macros are kept for future use, given the large extra-costs associated with this type of macros. In SOL-EP we allow an arbitrary (but still small) number of macros to be used in search, given the smaller extra-costs involved. SOL-EP macros have no preprocessing costs, and the cost per node in the search can be much smaller than in the case of CA-ED macros (see Table 4).

To decide the number of selected macros in a domain, a weight threshold $w_{im}$ is defined. This threshold can be seen as the weight of an imaginary macro $im$ with "constant performance" in all training instances. By "constant performance" we mean that, for each training instance,

$$\frac{N - N_{im}}{N} = c,$$

where $c > 0$ is a constant parameter. The threshold $w_{im}$ is updated following the same procedure as for regular macros: The initial value of $w_{im}$ is set to 1. For each training problem, the weight update of $im$ is

$$w_{im} = w_{im} - \alpha \delta_{im} L = w_{im} - \alpha \sigma(c) L$$

After all training problems have been processed, macros with a weight smaller than $w_{im}$ are selected for future use. In experiments we set $c$ to 0.01. Given the competition tight deadline, we invested limited time in studying this method and tuning its parameters. How to best determine the number of selected macros is still an open problem for us, which clearly needs more thourough study and evaluation.

### 3.3 Using Macros at Run-Time

The purpose of learned macros is to speed up search in new problem instances. A classical search algorithm expands a node by considering low-level actions that can be applied to the current state. We add successor states that can be reached by applying the whole sequence of actions in a macro. We order these macro successors before the regular successors of a state. Macros affects neither the completeness nor the correctness of the original algorithm. The completeness of an original search algorithm is preserved since SOL-EP removes no regular successors of a state. Correctness is guaranteed by the following way of applying a macro to a state. Given a state $s_0$ and a sequence of actions $m = a_1 a_2 ... a_k$ ($k = 2$ in our competition system), we say that $m$ is applicable to $s_0$ if $a_i$ can be applied to $s_{i-1}$, $i = 1, ..., k$, where $s_i = \gamma(s_{i-1}, a_i)$ and $\gamma(s, a)$ is the state obtained by applying $a$ to $s$.

When a given state is expanded at runtime, many instantiations of a macro could be applicable but only a few would actually be shortcuts towards a goal state. If all instantiations are considered, the branching factor can significantly increase and the induced overhead can be larger than the potential savings achieved by the useful instantiations. Therefore, the challenge is to select for state expansion only a small number of good macro instantiations. To determine what a "good" instantiation of a macro is, we use a heuristic method called *helpful macro pruning*. Helpful macro pruning is based on the relaxed graphplan computation that FF (Hoffmann & Nebel, 2001) performs for each evaluated state $s$. Given a state $s$, FF solves a relaxed problem, where the initial state is the currently evaluated state, the goal conditions are the same as in the real problem, and the actions are relaxed by ignoring their delete effects. This computation produces a relaxed plan $RP(s)$. In FF, the relaxed plan is used to heuristically evaluate problem states and prune low-level actions in the search space (helpful action pruning).

In addition, we use the relaxed plan to prune the set of macro-instantiations that will be used for node expansion. Since actions from the relaxed plan are often useful in the real world, we request that a selected macro and the relaxed plan match i.e., each action of the macro is part of the relaxed plan.

### 3.4 Discussion

The first part of this section summarizes properties of CA-ED macros and SOL-EP macros. Then comments on macro-ranking are provided, including a brief comparison of frequency-based ranking and gradient-descent ranking.

#### 3.4.1 CA-ED Macros vs SOL-EP Macros

When treated as single moves, macro-actions have the potential to influence the planning process in two important ways. First, macros can change the search space (the *embedding effect*), adding to a node successor list states that would normally be achieved in several steps. Intermediate states in the macro sequence do not have to be evaluated, reducing the search costs considerably. In effect, the maximal depth of a search could be reduced for the price of increasing the branching factor.

Second, macros can improve the heuristic evaluation of states (the *evaluation effect*). As shown before, FF computes this heuristic by solving a relaxed planning problem (i.e., the delete effects of actions are ignored) in a graphplan framework. To illustrate the benefits of macros in relaxed graphplan, consider the example in Figures 6 and 7. Operator UNLOAD has one add effect (LIFTING) and one delete effect (AVAILABLE) that update the status of a hoist from available (free) to lifting (busy). Similarly, operator DROP updates the hoist status with two such effects. However, when macro UNLOAD—DROP is used, the status of the hoist does not change: it was available (free) before, it will be available after. No effects are necessary to express changes in the hoist status. Hence two delete effects (one for each operator) are safely eliminated from the real problem before relaxation is performed. The relaxed problem is more similar to the real problem and the information loss is less drastic. See Section 5.4 for an empirical evaluation of how macros added to a domain affect the heuristic state evaluation with relaxed graphplan.

When macros can be added to the original domain formulation, both the evaluation effect and the embedding effect are present, with no need to extend the original planning engine. The disadvantages of this alternative include the limitation to STRIPS domains and, often, a significant increase of the preprocessing costs, memory requirements, or cost per node in the search (as shown in Section 5). When SOL-EP macros are used, each of the two effects needs a special extension of the planning engine. The current implementation of the enhanced planner handles the embedding effects but does not affect the computation of the heuristic state evaluation. Improving the heuristic state evaluation with macros is an important topic for future work.

### 3.4.2 Comments on Ranking

The frequency-based ranking method used with CA-ED is simple, fast and was shown to produce useful macros. Part of its success is due to the combination with static pruning rules. In particular, limiting macro length to only two actions simplifies the problem of macro ranking and filtering.

However, in the general case, the savings that a macro can achieve depend not only on how often it occurs as part of a solution, but also on several other factors, which can interact. Examples of such factors include the number of search nodes that the application of a macro would save, and the ratio of useful instantiations of a macro (providing shortcuts towards a goal state) versus instantiations that guide the search into a wrong direction. See the work of McCluskey and Porteous (1997) for more details on factors that determine the performance of macro-operators in AI planning.

The reason why we have extended our approach from frequency-based ranking to gradient-descent ranking is that integrating such factors as above into a ranking method is expected to produce more accurate results. Compared to frequency-based ranking, gradient-descent ranking measures the search performance of a macro more directly. To illustrate this, consider the solution plan in Figure 12. Table 2 shows the 5 distinct macro-operators extracted from this solution plan. For each macro, both the gradient-descent weight and the frequency-based weight are shown. In the latter case, the bonus points are ignored, since they do not affect the ranking (all macros will receive the same amount of bonus points for being part of this solution plan). Each method produces a different ranking. For example, macro TAKE-IMAGE TURN-TO is ranked fourth with the gradient-descent method and second with the frequency-based method. The reason is that a macro such as TURN-TO CALIBRATE (or SWITCH-ON TURN-TO) saves more search nodes than TAKE-IMAGE TURN-TO, even though it appears less frequently in the solution.

When compared to the simple and fast frequency-based method, gradient-descent ranking is more expensive. Each training problem has to be solved several times; once with no macros in use and once for each macro. As shown in Table 3 in Section 5, the training time can become an issue in domains such as PSR and Pipesworld Non-Temporal Tankage.

| Macro | Weight | Occurrences |
|---|---|---|
| turn-to take-image | 0.999103 | 3 |
| turn-to calibrate | 0.999103 | 1 |
| switch-on turn-to | 0.999103 | 1 |
| take-image turn-to | 0.999401 | 2 |
| calibrate turn-to | 0.999700 | 1 |

Table 2: Macros generated in the Satellite example.

Both ranking techniques ignore elements such as the interactions of several macros when used simultaneously, or the effects of macros on the quality of plans. See Section 5 for an evaluation of the latter.

Macro ranking is a difficult problem. The training data is often limited. In addition, factors such as frequency, number of search nodes that a macro would save, effects on solution quality, etc. have to be combined into a total ordering of a macro set. There is no clear best solution for this problem. For example, should we select a macro that speeds up planning but increases the solution length?

## 4. Implementation Enhancements in Macro-FF

This section describes the implementation enhancements added to FF with the goal of improving CPU and memory requirements. FF version 2.3 is highly optimized with respect to relaxed graphplan generation, which was assumed to be the performance bottleneck by the original system designers. We found that in several domains of the planning competition this assumption does not hold and the planner spends a significant portion of its time in other parts of the program. We applied two implementation enhancements to FF to reduce the CPU time requirements.

Another problem was that the memory requirements for some data structures built during the pre-processing stage grew exponentially with problem size and therefore did not scale. We replaced one of these data structures and were able to solve a few more problems in several domains within the memory limit used in the planning competition.

The enhancements described in this section affect neither the number of expanded nodes nor the quality of plans found by FF.

### 4.1 Open Queue

FF tries to find a solution using an enhanced hill climbing method and, if no solution is found, switches to a best-first search algorithm. Profiling runs showed that in the Pipesworld domains of the planning competition up to 90% of the CPU time is spent inserting nodes into the open queue. The open queue was implemented as a single linked list. We changed the implementation to use a linked list of buckets, one bucket for each heuristic value. The buckets are implemented as linked lists and need constant time for insertion, since they no longer have to be sorted.

### 4.2 State Hashing

The original FF already used state hashing to help identify previously visited states, with a full comparison of states in case of a collision. Each fact of a planning problem is assigned a unique 32-bit random number, and the hash code of a problem state is the sum of all random numbers associated to the facts that characterize the given state. Profiling runs showed that in some domains up to 35% of the CPU time is spent in the comparison of states. These are in particular domains with large states and small graphplan structures such as PSR and Philosophers. We replaced the original hash key by a 64-bit Zobrist hash key implementation, a standard technique in game-tree search (Marsland, 1986). Each fact is assigned 64-bit random number, and the hash key of a state is obtained applying the XOR operator to the random numbers corresponding to all facts true in the state.

When checking if two states are identical, only their hash codes are compared. If the hash codes are different, than the states are guaranteed to be different too. If the two compared states have the same hash code, we assume that the states are identical. This choice gives up completeness of a search algorithm: two different states with the same hash code can exist. However, this is so unlikely to occur that fast state comparison based on 64-bit Zobrist hashing is a common standard in high-performance game-playing programs. The large size of the hash key and the better randomization makes the occurrence of hash collisions much less probable than random hardware errors.

### 4.3 Memory Requirements

Some of the optimizations in FF require the creation of large lookup tables built during the preprocessing stage. One of them is a lookup table storing the facts of the initial state. This table is sparsely populated but the required memory is equal to the number of constants to the power of the arity of each predicate summed over all predicates in the domain. This caused the planner to run out of memory in some large domains given the 1 GB memory limit used in the planning competition. We replaced the lookup table by a balanced binary tree with minimal memory requirement and a lookup time proportional to the logarithm of the number of facts in the initial state.

### 5. Experimental Results

This section summarizes our experiments and analysis of results. We evaluate our ideas with several experiments, described in the next subsections. Section 5.1 evaluates the impact of the implementation enhancements on the planner's performance. Section 5.2 focuses on the effect of macro-operators in the system used in the competition. In these two subsections, the benchmarks that we competed in as part of IPC-4 are used for experimental evaluation: Promela Dining Philosophers – ADL (containing a total of 48 problems), Promela Optical Telegraph – ADL (48 problems), Satellite – STRIPS (36 problems), PSR Middle Compiled – ADL (50 problems), Pipesworld Notankage Nontemporal – STRIPS (50 problems), Pipesworld Tankage Nontemporal – STRIPS (50 problems), and Airport – ADL (50 problems). Macro-FF took the first place in Promela Optical Telegraph, PSR, and Satellite.

Section 5.3 compares the two abstraction techniques discussed in this article using STRIPS domains with static facts. We provide more details later in this section. Section 5.4 contains an empirical analysis of how CA-ED macros affect heuristic state evaluation and depth of goal states. All experiments reported in this article were run on a AMD Athlon 2 GHz machine, with the limits of 30 minutes and 1 GB of memory for each problem.

### 5.1 Enhanced FF

The new open queue implementation shows a significant speed-up in the Pipesworld domains. Figure 14 shows the difference in CPU time for the two different Pipesworld domains (note the logarithmic time scale). The simplest problems at the left of these charts are solved so quickly that no data bar is drawn. The speedup depends on the problem instance with maximum gains reaching a factor of 10. As a result two more problems were solved in the Pipesworld Tankage Non-Temporal domain and one more problem in the Pipesworld No-Tankage Non-Temporal domain.

The new 64-bit state hashing is especially effective in the PSR and Promela Dining Philosophers domains. Figure 15 shows a speed-up of up to a factor of 2.5. This resulted in 3 more problems solved in PSR, contributing to the success of Macro-FF in this domain.

The reduced memory requirement is important in Promela Optical Telegraph. Figure 16 shows the memory requirement of the original FF for the initial facts lookup table. As a result of the replacement of the lookup table, 3 more problems were solved in this domain.

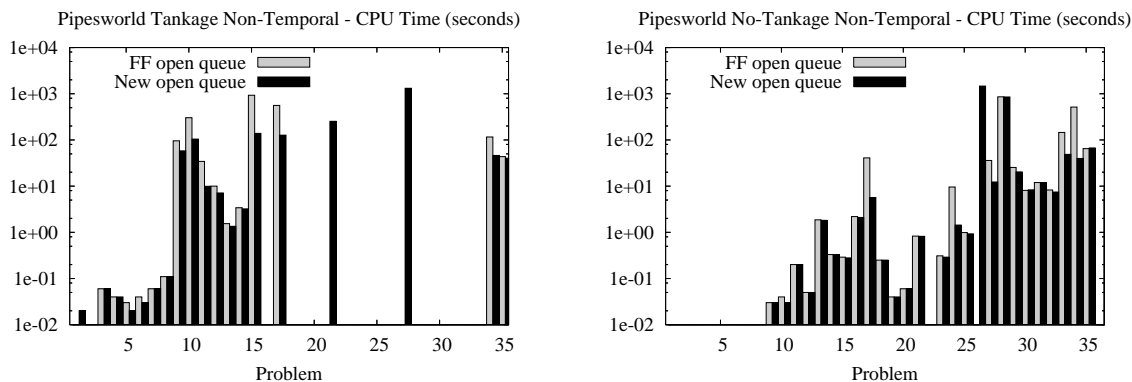

Figure 14: Comparison of old and new open queue implementation in Pipesworld Tankage Non-Temporal (left) and Pipesworld No-Tankage Non-Temporal (right). Results are shown for sets of 50 problems in each domain.

### 5.2 Evaluating Macros in the Competition System

In this subsection we evaluate how SOL-EP macros can improve performance in the competition system. We compare the planner with implementation enhancements against the planner with both implementation enhancements and SOL-EP macros.

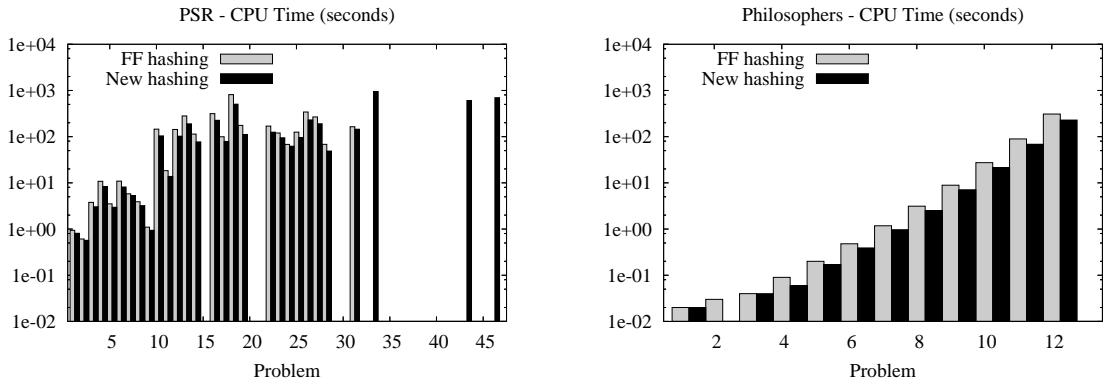

Figure 15: Comparison of the two implementations of state hashing in PSR (left) and Promela Dining Philosophers (right). Results are shown for 50 problems in PSR and 48 problems in Promela Dining Philosophers.

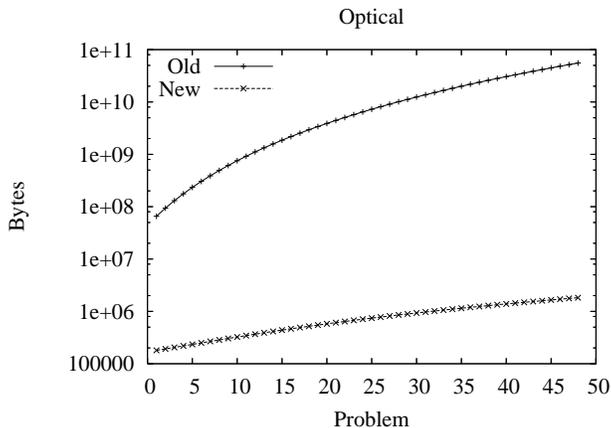

Figure 16: Size of the data structures for initial facts in the old implementation (lookup table) and the new implementation (balanced tree) in Promela Optical Telegraph.

For each of the seven test domains, we show the number of expanded nodes and the total CPU time, again, on a logarithmic scale. A CPU time chart shows no distinction between a problem solved very quickly (within a time close to 0) and a problem that could not be solved. To determine what the case is, check the corresponding node chart, where the absence of a data point always means no solution.

Figure 17 summarizes the results in Satellite, Promela Optical Telegraph, and Promela Dining Philosophers. In Satellite and Promela Optical Telegraph, macros greatly improve performance over the whole problem sets, allowing Macro-FF to win these domain formu-Botea, Enzenberger, Müller, & Schaefferlations in the competition. In Promela Optical Telegraph macros led to solving 12 additional problems. The savings in Promela Dining Philosophers are limited, resulting in one more problem solved.

Figure 18 shows the results in the ADL version of Airport. The savings in terms of expanded nodes are significant, but they have little effect on the total running time. In this domain, the preprocessing costs dominate the total running time.

The complexity of preprocessing in Airport also limits the number of solved problems to 21. The planner can solve more problems when the STRIPS version of Airport is used, but no macros could be generated for this domain version. STRIPS Airport contains one domain definition for each problem instance, whereas our learning method requires several training problems for a domain definition.

Figure 19 contains the results in Pipesworld Non-Temporal No-Tankage, Pipesworld Non-Temporal Tankage, and PSR. In Pipesworld Non-Temporal No-Tankage, macros often lead to significant speed-up. As a result, the system solves four new problems. On the other hand, the system with macros fails in three previously solved problems. The contribution of macros is less significant in Pipesworld Non-Temporal Tankage. The system with macros solves two new problems and fails in one previously solved instance. Out of all seven benchmarks, PSR is the domain where macros have the smallest impact. Both systems solve 29 problems using similar amounts of resources. In the competition official run, Macro-FF solved 32 problems in this domain formulation.

Table 3 shows the number of training problems, the total training time, and the selected macros in each domain. The training phase uses 10 problems for each of Airport, Satellite, Pipesworld Non-Temporal No-Tankage, and PSR. We reduced the training set to 5 problems for Promela Optical Telegraph, 6 problems for Promela Dining Philosophers, and 5 problems for Pipesworld Non-Temporal Tankage. In Promela Optical Telegraph, the planner with no macros solves 13 problems, and using most of them for training would leave little room for evaluating the learned macros. The situation is similar in Promela Dining Philosophers; the planner with no macros solves 12 problems. In Pipesworld Non-Temporal Tankage, the smaller number of training problems is caused by both the long training time and the structure of the competition problem set. The first 10 problems use only a part of the domain operators, so we did not include these into the training set. Out of the remaining problems, the planner with no macros solves 11 instances. The large training times in Pipesworld Non-Temporal Tankage and PSR are caused by the increased difficulty of the training problems.

### 5.3 Evaluating our Abstraction Techniques

To evaluate the performance of our two abstraction methods, we compare four setups of Macro-FF. In all four setups, the planner includes the implementation enhancements described in Section 4. Setup 1 is the planner with no macros. Setup 2 includes CA-ED, the method described in Section 2. Setup 3 uses SOL-EP, the method described in Section 3. Setup 4 is a combination of 2 and 3. Since both methods have benefits and limitations, it is interesting to analyze how they perform when applied together. In setup 4, we first run CA-ED, obtaining an enhanced domain. Next we treat this as a regular domain, and

606

| Domain | TP | TT | Macros |
|---|---|---|---|
| Airport | 10 | 365 | MOVE MOVE <br> PUSHBACK MOVE <br> PUSHBACK PUSHBACK <br> MOVE TAKEOFF |
| Promela Optical Telegraph | 5 | 70 | QUEUE-WRITE ADVANCE-EMPTY-QUEUE-TAIL <br> ACTIVATE-TRANS QUEUE-WRITE <br> ACTIVATE-TRANS ACTIVATE-TRANS <br> PERFORM-TRANS ACTIVATE-TRANS |
| Promela Dining Philosophers | 6 | 10 | ACTIVATE-TRANS QUEUE-READ <br> ACTIVATE-TRANS ACTIVATE-TRANS <br> QUEUE-READ ADVANCE-QUEUE-HEAD |
| Satellite | 10 | 8 | TURN-TO SWITCH-ON <br> SWITCH-ON TURN-TO <br> SWITCH-ON CALIBRATE <br> TURN-TO TAKE-IMAGE <br> TURN-TO CALIBRATE <br> TAKE-IMAGE TURN-TO |
| Pipesworld Non-Temporal No-Tankage | 10 | 250 | POP-START POP-END <br> PUSH-START PUSH-END <br> PUSH-START POP-START |
| Pipesworld Non-Temporal Tankage | 5 | 4,206 | PUSH-START PUSH-END <br> PUSH-START POP-END <br> PUSH-END POP-START <br> POP-END PUSH-START <br> PUSH-END PUSH-START <br> PUSH-START POP-START <br> POP-START PUSH-START |
| PSR | 10 | 1,592 | AXIOM AXIOM <br> CLOSE AXIOM |

Table 3: Summary of training in each domain. TP is the number of training problems and TT is the total training time in seconds. The last column shows the macros selected for each domain. For simplicity, we do not show the variable mapping of each macro.

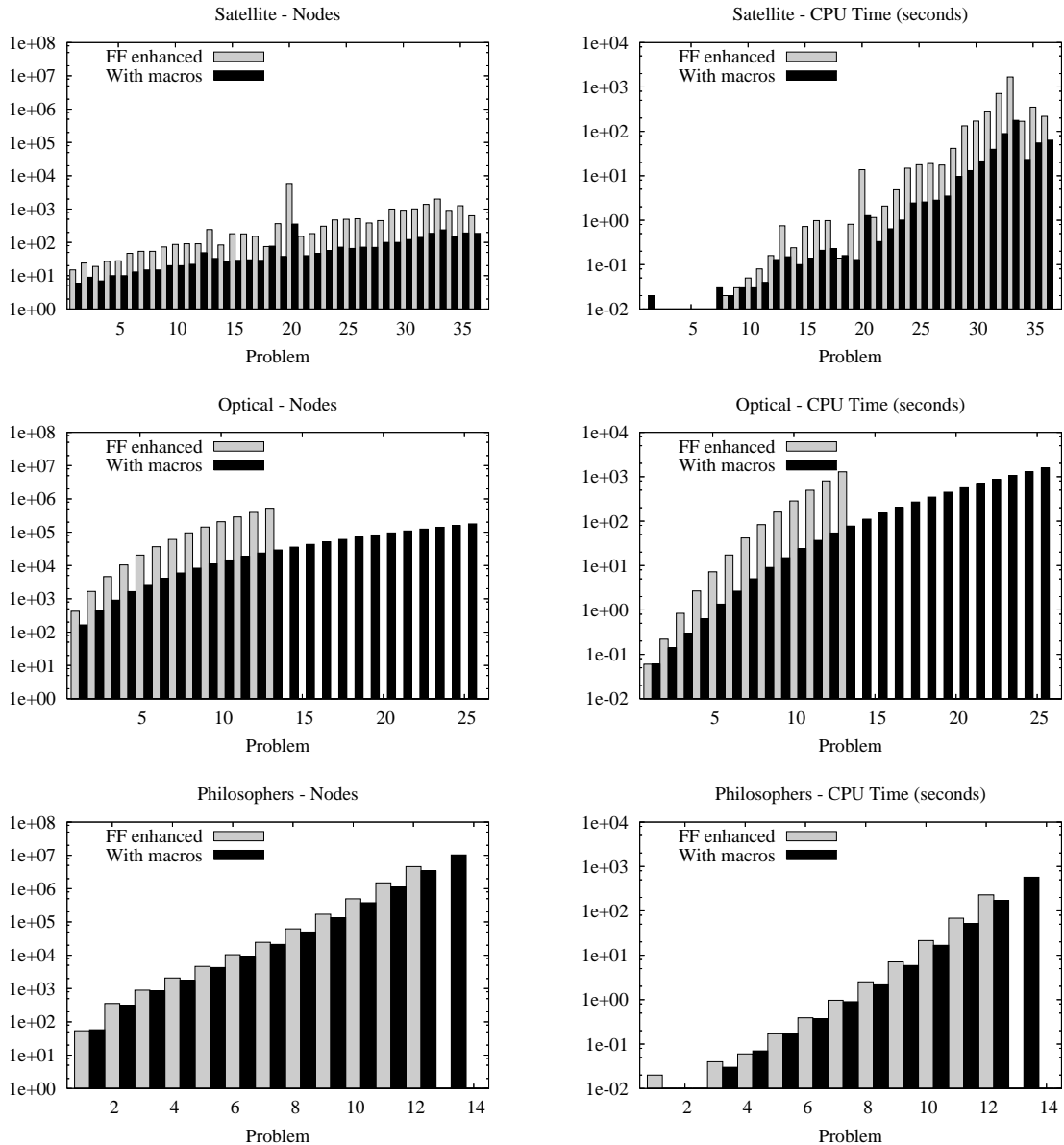

Figure 17: Comparison of our enhanced version of FF with and without macros in Satellite (36 problems), Promela Optical Telegraph (48 problems) and Promela Dining Philosophers (48 problems).

apply SOL-EP to generate a list of SOL-EP macros. Finally, the enhanced planner uses as input the enhanced domain, the list of SOL-EP macros, and regular problem instances.

Since component abstraction can currently be applied only to STRIPS domains with static facts in their formulation, we used as testbeds domains that satisfy these constraints.

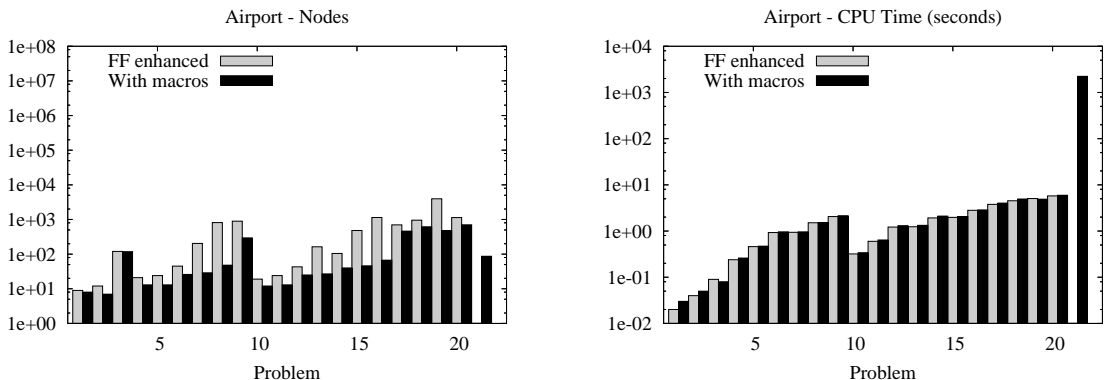

Figure 18: Comparison of our enhanced version of FF with and without macros in Airport (50 problems in total).

We ran this experiment on Rovers (20 problems), Depots (22 problems), and Satellite (36 problems). These domains were used in the third international planning competition IPC-3, and Satellite was re-used in IPC-4 with an extended problem set. In our experiments, the Rovers and Depots problems sets are the same as in IPC-3, and the Satellite problem set is the same as in IPC-4.

Figures 20 – 23 and Table 4 summarize our results. The performance consistently improves when macros are used. Interestingly, combining CA-ED and SOL-EP often leads to better performance than each abstraction method taken separately. In Rovers, all three abstraction setups produce quite similar results, with a slight plus for the combined setup. In Depots, CA-ED is more effective than SOL-EP in terms of expanded nodes. The differences in CPU time become smaller, since adding new operators to the original domain significantly increases the cost per node in Depots (see the discussion below). Again, the overall winner in this domain is the combined setup. In Satellite, adding macros to the domain reduces the number of expanded nodes, but has significant impact in cost per node (see Table 4 later in this section) and memory requirements. Setups 2 and 4, which add macros to the original domain, fail to solve three problems (32, 33, and 36) because of the large memory requirements in FF's preprocessing phase.

Table 4 evaluates how macros can affect the *cost per node* in the search. The cost per node is defined as the total search time divided by the number of evaluated states. A value in the table is the cost per node in the corresponding setup (i.e., CA-ED or SOL-EP) divided by the cost per node in the setup with no macros. For each of the two methods we show the minimum, the maximum, and the average value. When macros are added to the original domain (i.e., the domain is enhanced), the increase in cost per node can be significant. The average rate is 7.70 in Satellite, and 6.06 in Depots. It is interesting to notice that this cost is less than 1 in Rovers. This is an effect of solving a problem with less nodes expanded. Operations such as managing the open list and the closed list have costs that increase with the size of the lists at a rate that can be higher than linear. The right part of the table shows much better values for the cost rate when macros are used in an enhanced planner.

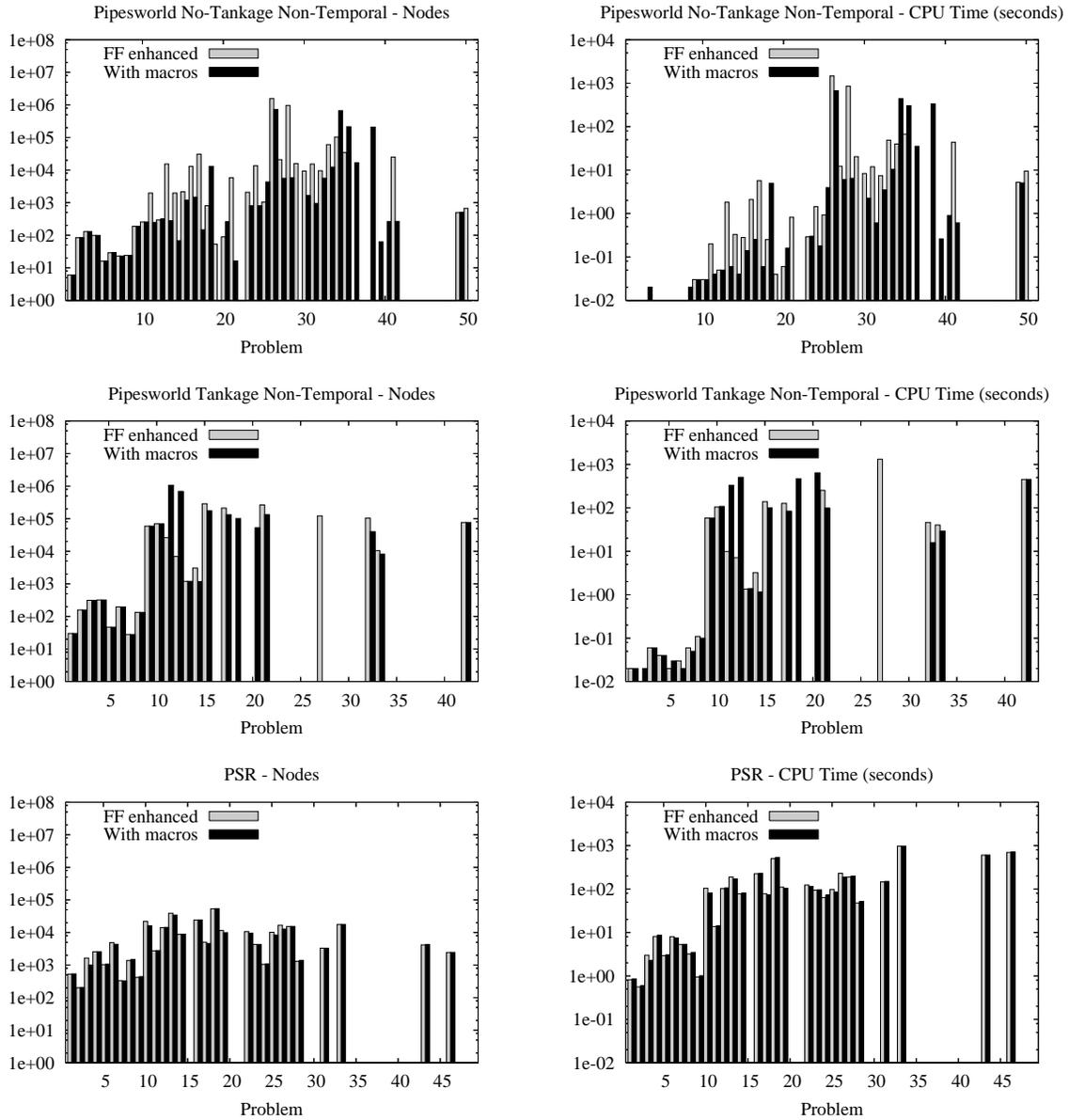

Figure 19: Comparison of our enhanced version of FF with and without macros in Pipesworld No-Tankage Non-Temporal, Pipesworld Tankage Non-Temporal and PSR (50 problems for each domain).

It is important to analyze why macros added as new operators generate such an increase in cost per node in Satellite and Depots. The overhead is mostly present in the relaxed graphplan algorithm that computes the heuristic value of a state. The complexity of this algorithm depends upon the total number of actions that have been instantiated during preprocessing for a given problem. Adding new operators to a domain increases the number of pre-instantiated actions. Since macros tend to have more variables than a regular operator, the corresponding number of instantiations can be significantly larger. Let the *action instantiation rate* be the number of actions instantiated for a problem when macros are used divided by the number of actions instantiated in the original domain formulation. Our statistics show that the average action instantiation rate is 6.03 in Satellite, 3.20 in Depots, and 1.04 in Rovers.

The results show no significant impact of macro-operators on the solution quality. When macros are used, the length of a plan slightly varies in both directions, with an average close to the value of the original FF system.

| Domain | CA-ED | | | SOL-EP | | |
|---|---|---|---|---|---|---|
| | Min | Max | Avg | Min | Max | Avg |
| Depots | 3.27 | 8.56 | 6.06 | 0.93 | 1.14 | 1.04 |
| Rovers | 0.70 | 0.90 | 0.83 | 0.85 | 1.14 | 1.00 |
| Satellite | 0.98 | 14.38 | 7.70 | 0.92 | 1.48 | 1.11 |

Table 4: Relative cost per node.

### 5.4 Evaluating the Effects of CA-ED Macros on Heuristic State Evaluation

As shown in Section 3.4, macros added to a domain as new operators affect both the structure of the search space (the embedding effect) and the heuristic evaluation of states with relaxed graphplan (the evaluation effect). This section presents an empirical analysis of these.

Figure 24 shows results for Depots, Rovers and Satellite. For each domain, the chart on the left shows data for the original domain formulation, and the chart on the right shows data for the macro-enhanced domain formulation. For each domain formulation, the data points are extracted from solution plans as follows. Each state along a solution plan generates one data point. The coordinates of the data point are the state's heuristic evaluation on the vertical axis, and the number of steps left until the goal state is reached on the horizontal axis. The number of steps to a goal state may be larger than the distance (i.e., length of shortest path) to a goal state. The reason why states along solution plans were used to generate data is that for such states, both the heuristic evaluation, and the number of steps to a goal state are available after solving a problem.

The first conclusion from Figure 24 is that macros added to a domain improve the accuracy of heuristic state evaluation of relaxed graphplan. The closer a data point is to the diagonal, the more accurate the heuristic evaluation of the corresponding state.

Secondly, data clouds are shorter in macro-enhanced domains. This is a direct result of the embedding effect, which reduces the depth of goal states.

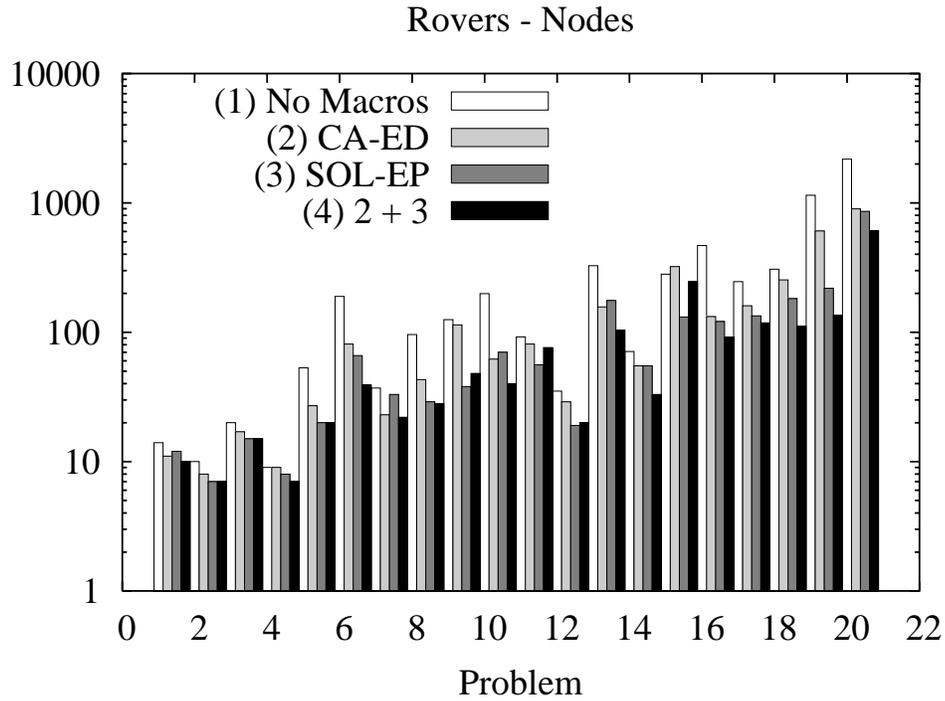

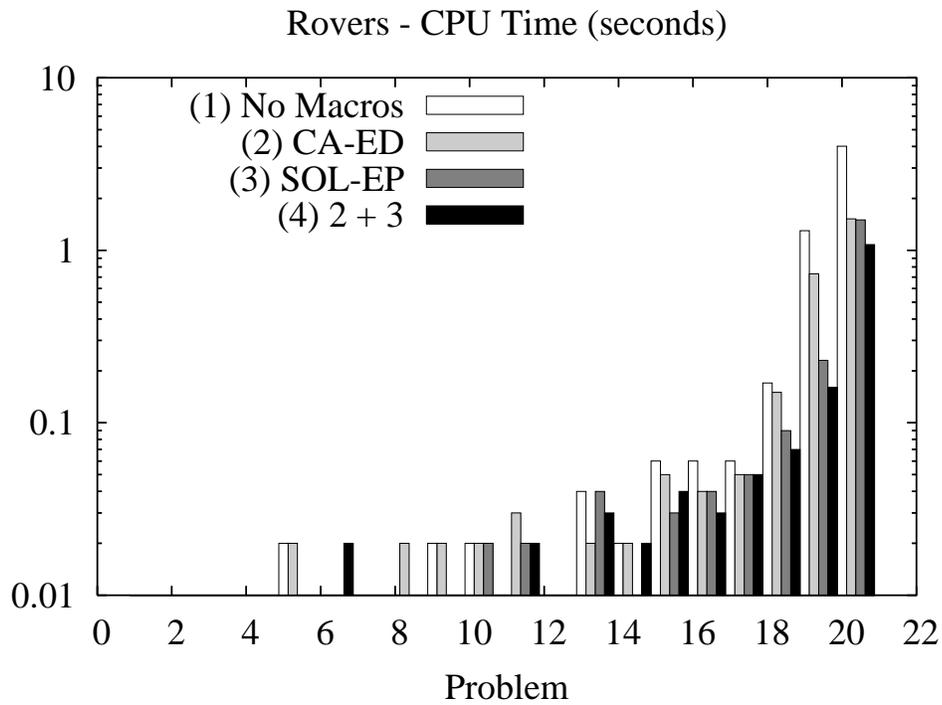

Figure 20: Evaluating abstraction techniques in Rovers. We show the number of expanded nodes (top), and the CPU time (bottom).

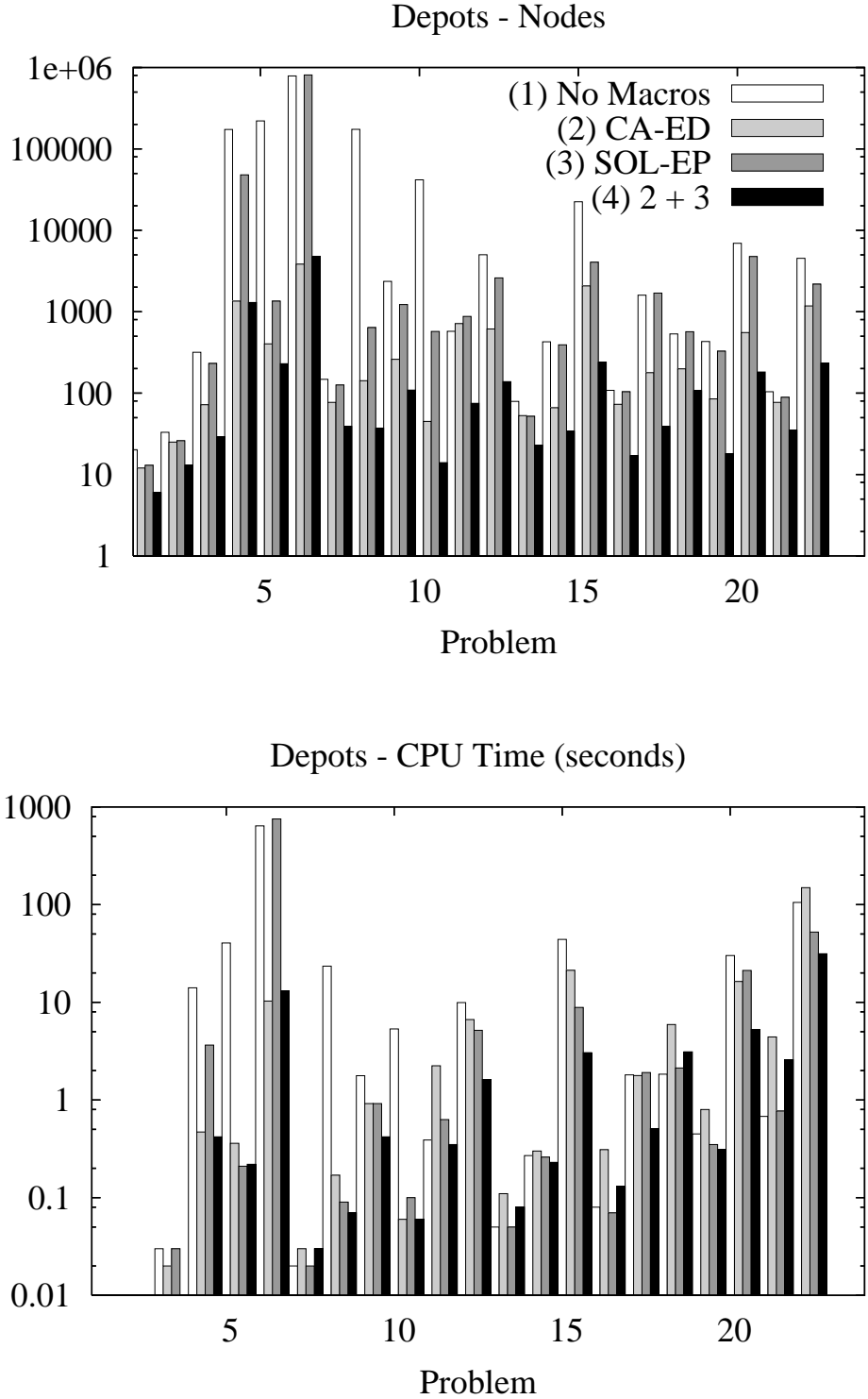

Figure 21: Evaluating abstraction techniques in Depots. We show the number of expanded nodes (top), and the CPU time (bottom).

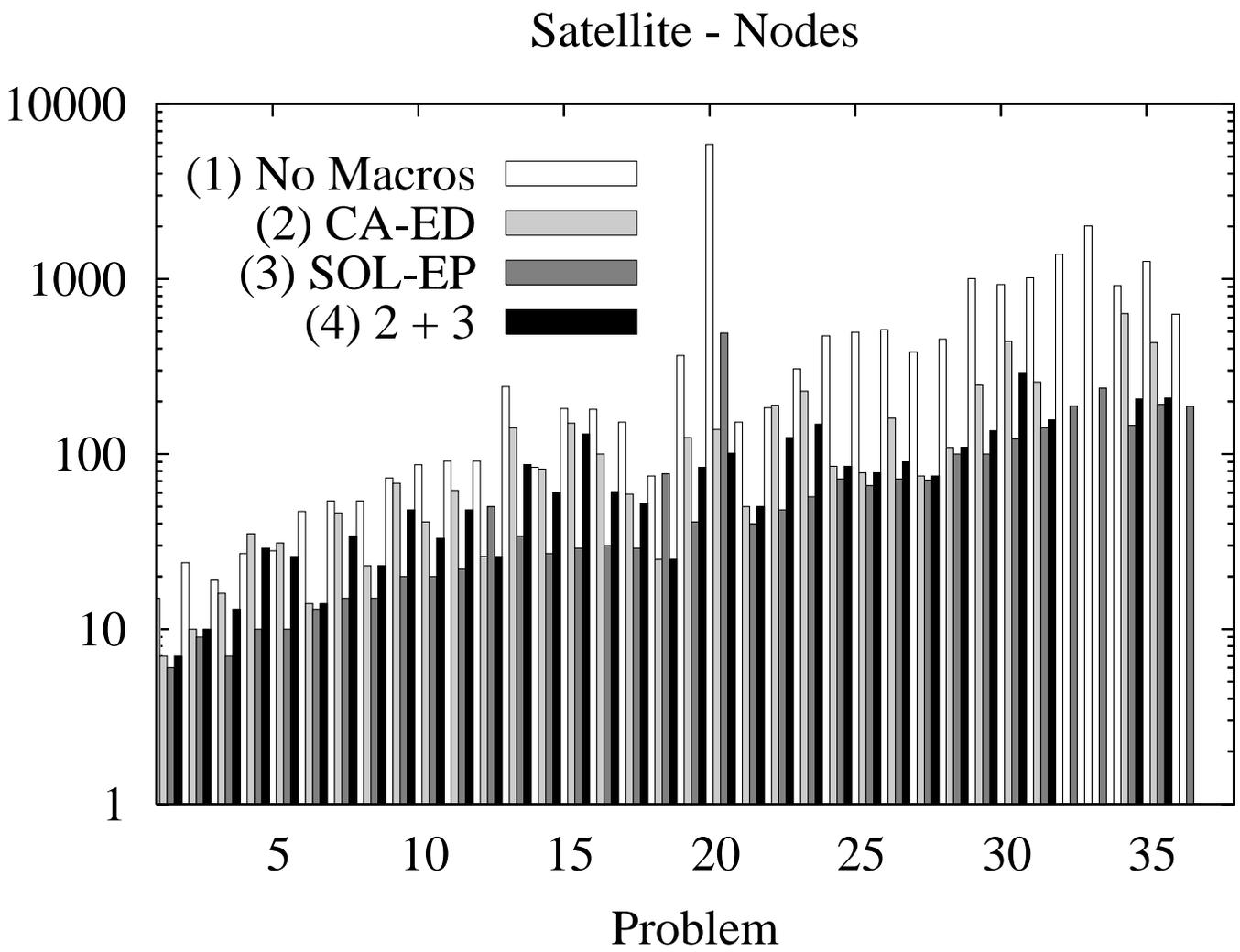

Figure 22: Evaluating abstraction techniques in Satellite. We show the number of expanded nodes.

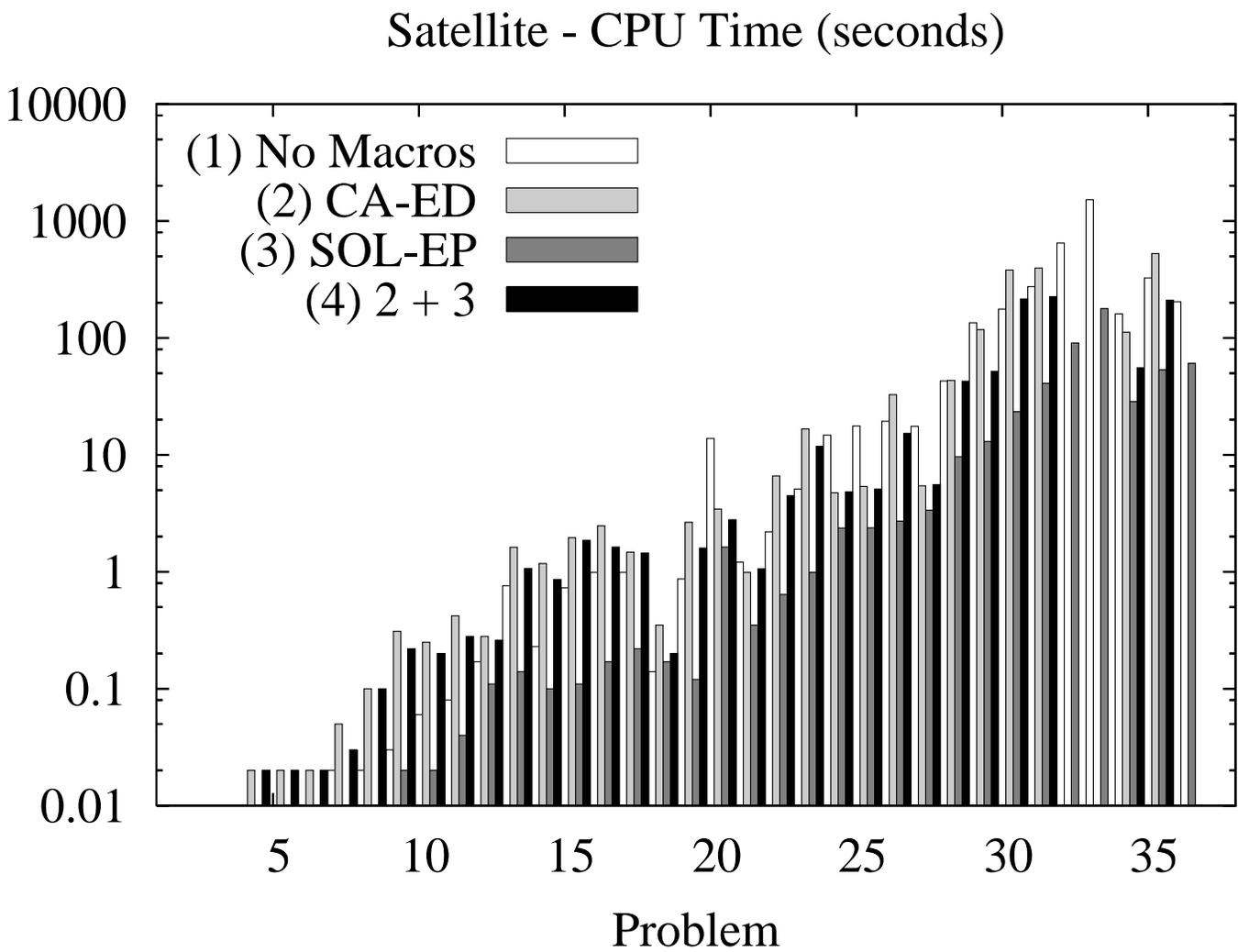

Figure 23: Evaluating abstraction techniques in Satellite (continued). We show the CPU time.

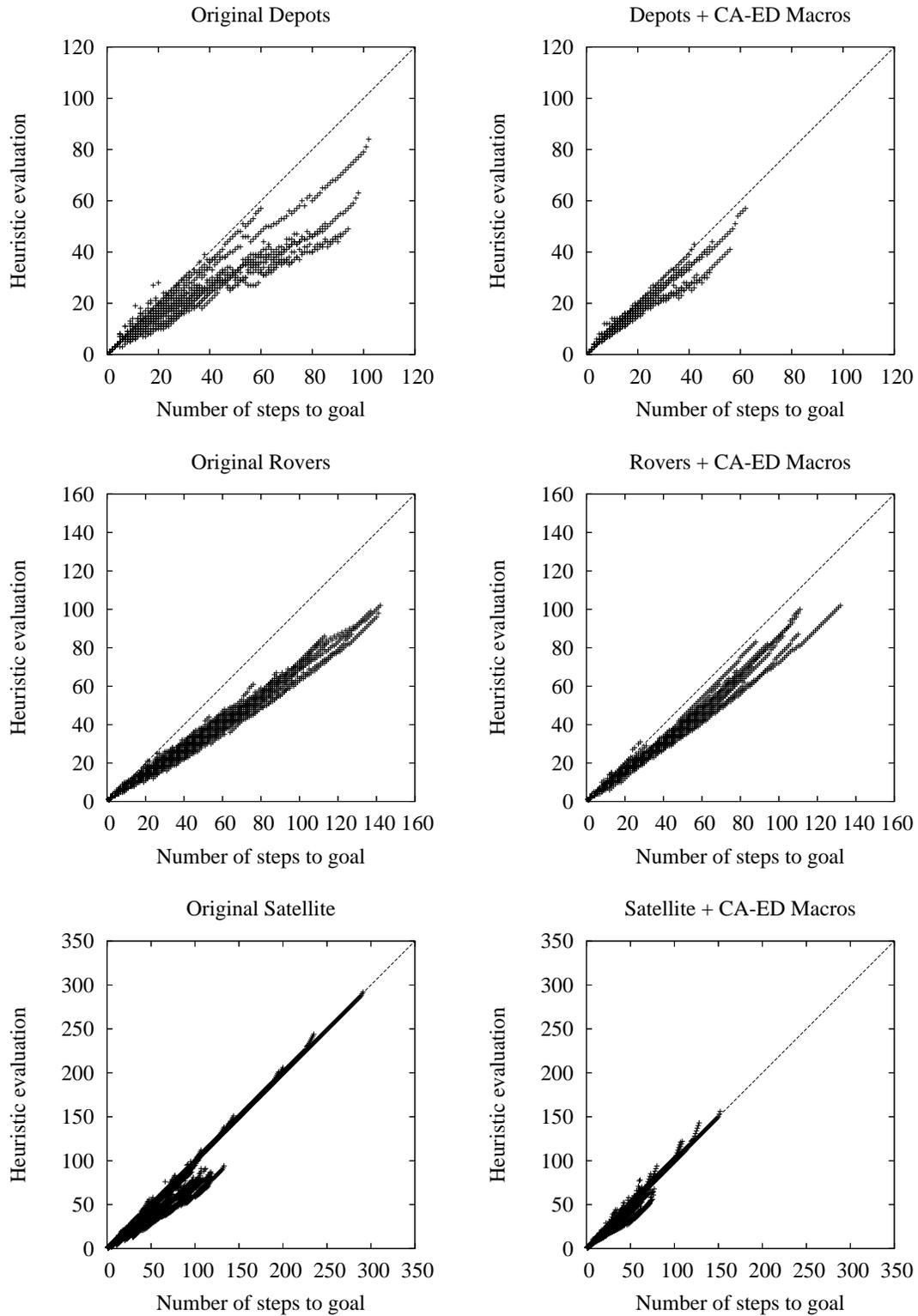

Figure 24: Effects of CA-ED macros on heuristic state evaluation and depth of goal states.

## 6. Related Work

The related work described in this section falls into two categories. We first review approaches that make use of the domain structure to reduce the complexity of planning, and next consider previous work on macro-operators.

An automatic method that discovers and exploits domain structure has been explored by Knoblock (1994). In this work, a hierarchy of abstractions is built starting from the initial low-level problem description. A new abstract level is obtained by dropping literals from the problem definition at the previous abstraction level. Planning first produces an abstract solution and then iteratively refines it to a low-level representation. The hierarchy is built in such a way that, if a refinement of an abstract solution exists, no backtracking across abstraction levels is necessary during the refinement process. Backtracking is performed only when an abstract plan has no refinement. Such situations can be arbitrarily frequent, with negative effects on the system's performance.

Bacchus and Yang (1994) define a theoretical probabilistic framework to planning in hierarchical models. Abstract solutions of a problem at different abstraction levels are hierarchically represented as nodes in a tree structure. A tree edge indicates that the target node is a refinement of the start node. An abstract solution can be refined to the previous level with a given probability. Hierarchical search in this model is analytically evaluated. The analytical model is further used to enhace Knoblock's abstraction algorithm. The enhancement refers to using estimations of the refinement probabilities for abstract solutions.

More recently, implicit causal structure of a domain has been used to design a domain-independent heuristic for state evaluation (Helmert, 2004). These methods either statically infer information about the structure of a domain, or dynamically discover the structure for each problem instance. In contrast, we propose an adaptive technique that learns from previous experience in a domain.

Two successful approaches that use hand-crafted information about the domain structure are hierarchical task networks and temporal logic control rules. Hierarchical task networks (HTNs) guide and restrict planning by using a hierarchical representation of a domain. Human experts design hierarchies of tasks that show how the initial problem can be broken down to the level of regular actions. The idea was introduced by Sacerdoti (1975) and Tate (1977), and has widely been used in real-life planning applications (Wilkins & desJardins, 2001). SHOP2 (Nau, Au, Ilghami, Kuter, Murdock, Wu, & Yaman, 2003) is a well-known heuristic search planner where search is guided by HTNs.

In planning with temporal logic control rules, a formula is associated with each state in the problem space. The formula of the initial state is provided with the domain description. The formula of any other state is obtained based on its successor's formula. When the formula associated with a state can be proven false, that state's subtree is pruned. The best known planners of this kind are TLPlan (Bacchus & Kabanza, 2000) and TALPlanner (Kvarnström & Doherty, 2001). While efficient, these approaches also rely heavily on human knowledge, which might be expensive or impossible to obtain.

Early contributions to macro-operators in AI planning includes the work of Fikes and Nilsson (1971). Macros are extracted after a problem was solved and the solution became available. Minton (1985) advances this work by introducing techniques that filter the set of learned macro-operators. In his approach, two types of macro-operators are preferred: S-macros, which occur with high frequency in problem solutions, and T-macros, which can be useful but have low priority in the original search algorithm. Iba (1989) introduces the so-called peak-to-peak heuristic to generate macro-operators at run-time. A macro is a move sequence between two peaks of the heuristic state evaluation. Such a macro traverses a "valley" in the search space, and using this later can correct errors in the heuristic evaluation. Mooney (1988) considers whole plans as macros and introduces partial ordering of operators based on their causal interactions.

Veloso and Carbonell (1993) and Kambhampati (1993) explore how planning can reuse solutions of previously solved problems. Solutions annotated with additional relevant information are stored for later use. This additional information contains either explanations of successful or failed search decisions (Veloso & Carbonell, 1993), or the causal structure of solution plans (Kambhampati, 1993). Several similarity metrics for planning problems are introduced. When a new problem is fed to the planner, the annotated solutions of similar problems are used to guide the current planning process.

McCluskey and Porteous (1997) focus on constructing planning domains starting from a natural language description. The approach combines human expertise and automatic tools, and addresses both correctness and efficiency of the obtained formulation. Using macro-operators is a major technique that the authors propose for efficiency improvement. In this work, a state in a domain is composed of local states of several variables called dynamic objects. Macros model transitions between the local states of a variable.

The planner Marvin (Coles & Smith, 2004) generates macros both online (as plateau-escaping sequences) and offline (from a reduced version of the problem to be solved). No macros are cached from one problem instance to another.

Macro-moves were successfully used in single-agent search problems such as puzzles or path-finding in commercial computer games, usually in a domain-specific implementation. The sliding-tile puzzle was among the first testbeds for this idea (Korf, 1985; Iba, 1989). Two of the most effective concepts used in the Sokoban solver Rolling Stone, tunnel and goal macros, are applications of this idea (Junghanns & Schaeffer, 2001). More recent work in Sokoban includes an approach that decomposes a maze into a set of rooms connected by tunnels (Botea, Müller, & Schaeffer, 2002). Search is performed at the higher level of abstract move sequences that rearrange the stones inside a room so that a stone can be transferred from one room to another. Using macro-moves for solving Rubik's Cube puzzles is proposed by Hernádvölgyi (2001). A method proposed by Botea, Müller, and Schaeffer (2004a) automatically decomposes a navigation map into a set of clusters, possibly on several abstraction levels. For each cluster, an internal optimal path is pre-computed between any two entrances of that cluster. Path-finding is performed at an abstract level, where a macro-move crosses a cluster from one entrance to another in one step.

Methods that exploit at search time the relaxed graphplan associated with a problem state (Hoffmann & Nebel, 2001) include helpful action pruning (Hoffmann & Nebel, 2001) and look-ahead policies (Vidal, 2004). Helpful action pruning considers for node expansion only actions that occur in the relaxed plan and can be applied to the current state. Helpful macro pruning applies the same pruning idea for the macro-actions applicable to a state, with the noticeable difference that helpful macro pruning does not give up completeness of the search algorithm. A lookahead policy executes parts of the relaxed plan in the real world, as this often provides a path towards a goal state with no search and few states evaluated. The actions in the relaxed plan are iteratively applied as long as this is possible in a heuristically computed order. When the lookahead procedure cannot be continued with actions from the relaxed plan, a plan-repair method selects a new action to be applied.

## 7. Conclusion and Future Work

Despite the great progress that AI planning has recently achieved, many benchmarks remain challenging for current planners. In this paper we presented techniques that automatically learn macro-operators from previous experience in a domain, and use them to speed up the search in future problems. We evaluated our methods on standard benchmarks from international planning competitions, showing significant improvement for domains where structure information can be inferred. We implemented our ideas in Macro-FF, an extension of FF version 2.3. Macro-FF participated in the classical part of the fourth international planning competition, competing in seven domains and taking first place in three of them.

Exploring our method more deeply and improving the performance in more classes of problems are major directions for future work. We also plan to extend our approach in several directions. Our learning method can be generalized from macro-operators to more complex structures such as HTNs. Little research focusing on learning HTNs has been conducted, even though the problem is of great importance.

We plan to explore how a heuristic evaluation based on the relaxed graphplan can be improved with macro-operators. As shown in this article, a macro added to an original domain formulation as a regular operator influences the results of the heuristic function. This is convenient (no changes are necessary in the planning engine), but limited only to STRIPS domains. For other subsets of PDDL, the relaxed graphplan algorithm can be extended with special capabilities to handle macros when no enhanced domain definition is provided.

The long-term goal of component abstraction is automatic reformulation of planning domains and problems. When a real-world problem is abstracted into a planning model, the corresponding formulation is expressed at an abstraction level that a human designer considers appropriate. However, choosing a good abstraction level could be a hard and expensive problem for humans. Hence methods that automatically update the formulation of a problem based on its structure would be helpful.

## Acknowledgments

This research was supported by the Natural Sciences and Engineering Research Council of Canada (NSERC) and Alberta's Informatics Circle of Research Excellence (iCORE). We thank Jörg Hoffmann for making the source code of FF available, and for kindly answering many technical questions related to FF. We also thank the organizers of IPC-4, the reviewers of this article, and Maria Fox, who led the reviewing process.